\documentclass{article}

\usepackage{microtype}
\usepackage{graphicx}
\usepackage{subfigure}
\usepackage{booktabs} 
\usepackage{url}
\usepackage{pifont}
\usepackage{balance}
\usepackage{hyperref}
\usepackage{bigstrut,multirow,rotating,booktabs}
\usepackage{array}
\usepackage[table]{xcolor}
\usepackage{multirow}
\usepackage{bigstrut}
\usepackage{amsmath}
\usepackage{colortbl}
\usepackage{caption} 
\usepackage{setspace}
\usepackage{makecell}
\usepackage{enumitem}
\usepackage{amssymb}
\usepackage{algorithm}
\usepackage{algorithmic}
\usepackage{listings}
\usepackage{tcolorbox}

\definecolor{mybrown1}{RGB}{194, 154, 119}
\definecolor{mybrown2}{RGB}{131, 77, 89}
\definecolor{mybrown3}{RGB}{96, 31, 59}

\definecolor{mygray}{RGB}{220,220,220}
\definecolor{mygray1}{RGB}{230,230,230}
\definecolor{mygray2}{RGB}{240,240,240}

\usepackage{hyperref}

\usepackage[accepted]{icml2025}

\usepackage{amsmath}
\usepackage{amssymb}
\usepackage{mathtools}
\usepackage{amsthm}

\usepackage[capitalize,noabbrev]{cleveref}

\theoremstyle{plain}

\theoremstyle{definition}

\theoremstyle{remark}

\usepackage[textsize=tiny]{todonotes}

\icmltitlerunning{OpenDDI: A Comprehensive Benchmark for DDI Prediction}

\begin{document}

\twocolumn[
\icmltitle{OpenDDI: A Comprehensive Benchmark for DDI Prediction}



\icmlsetsymbol{equal}{*}

\begin{icmlauthorlist}
\icmlauthor{Xinmo Jin}{equal}
\icmlauthor{Bowen Fan}{equal}
\icmlauthor{Xunkai Li}{equal}
\icmlauthor{Henan Sun}{}
\icmlauthor{YuXin Zeng}{} 
\icmlauthor{Zekai Chen}{}
\icmlauthor{Yuxuan Sun}{}
\icmlauthor{Jia Li}{}
\icmlauthor{Qiangqiang Dai}{}
\icmlauthor{Hongchao Qin}{}
\icmlauthor{Rong-Hua Li}{}
\icmlauthor{Guoren Wang}{}

\end{icmlauthorlist}



\icmlcorrespondingauthor{Rong-Hua Li}{lironghuabit@126.com}

\icmlkeywords{Machine Learning, ICML}

\vskip 0.3in
]



\printAffiliationsAndNotice{\icmlEqualContribution} 

\begin{abstract}
Drug–Drug Interactions (DDIs) significantly influence therapeutic efficacy and patient safety. As experimental discovery is resource-intensive and time-consuming, efficient computational methodologies have become essential. The predominant paradigm formulates DDI prediction as a drug graph-based link prediction task.
However, further progress is hindered by two fundamental challenges: (1) lack of high-quality data: most studies rely on small-scale DDI datasets and single-modal drug representations; (2) lack of standardized evaluation: inconsistent scenarios, varied metrics, and diverse baselines.
To address the above issues, we propose OpenDDI, a comprehensive benchmark for DDI prediction.
Specifically, (1) from the data perspective, OpenDDI unifies 6 widely used DDI datasets and 2 existing forms of drug representation, while additionally contributing 3 new large-scale LLM-augmented datasets and a new multimodal drug representation covering 5 modalities.
(2) From the evaluation perspective, OpenDDI unifies 20 SOTA model baselines across 3 downstream tasks, with standardized protocols for data quality, effectiveness, generalization, robustness, and efficiency. 
Based on OpenDDI, we conduct a comprehensive evaluation and derive 11 valuable insights for DDI prediction while exposing current limitations to provide critical guidance for this rapidly evolving field.
Our code is available at \url{https://github.com/xiaoriwuguang/OpenDDI}

\end{abstract}

\vspace{-0.6cm}

\section{Introduction}
With the expansion of pharmacotherapy and the prevalence of polypharmacy, alterations in drug effects due to co-administration have emerged as a common phenomenon in medication. We formally refer to such phenomena as Drug–Drug Interactions (DDIs)~\cite{ddi1}, which may produce effects like enhanced therapeutic efficacy or aggravated disease progression ~\cite{risk1,risk2,risk3}. Therefore, identifying and understanding DDIs is of paramount importance~\cite{importance1,importance2}, which is critical to patient safety~\cite{patientsafe}, polypharmacy strategies~\cite{polypharmacy}, and drug development~\cite{advancingdrug}.

\begin{table*}[t]
\vspace{-0.3cm}
\caption{DDI prediction benchmark comparison.}
\centering
\fontsize{9}{8.5}\selectfont    
\renewcommand{\arraystretch}{1.3}  
\setlength{\tabcolsep}{4pt}        
\label{tab:DDI-benchmark-comparison}
\begin{tabular}{c|c|c|c|c|c}
\midrule[0.3pt]
\textbf{Benchmarks} & \textbf{Datasets} & \textbf{Representations} & \textbf{Algorithms} & \textbf{Tasks} & \textbf{Evaluations} \\ 
\midrule[0.3pt]
DDI-Ben~\cite{ddi-ben} & 2 small-scale & Single-modal & 
10 + & 
2 & 
Generalization, Robustness \\ 
\midrule[0.3pt]
OpenDDI (Ours) & 9 (3 million-level) & Multimodal & 
20 + & 
3 & 
\begin{tabular}[c]{@{}c@{}}Quality,  Effectiveness,Efficiency\\ Generalization, Robustness \end{tabular} \\ 
\midrule[0.3pt]
\end{tabular}
\vspace{-0.5cm}
\end{table*}
  
However, traditional laboratory-based approaches for validating DDIs often require significant time and extensive resources~\cite{labcost,labcost2}.
To address aforementioned issues, machine learning~\cite{machine-learning}, particularly graph-based approaches, offers a promising alternative by modeling drugs as nodes and interactions as edges, with their pharmacological attributes used as node features, which casts DDI prediction as a link prediction task effectively addressed by graph neural networks (GNNs)~\cite{gnn2,gnn1}. These methods have achieved strong performance while reducing experimental costs~\cite{gnn4ddi1,gnn4ddi2}.

Despite these advances, studies still remain constrained by two key challenges: 
(1) \textbf{Limitations in data quality}: Due to policy constraints and limited accessibility, DDI metadata remain difficult to collect, leading current methods to depend on small-scale datasets that usually contain only few drug nodes and sparse DDI edges. Besides, most methods rely on single-modal drug representations such as SMILES or path, which primarily capture structural or relational information while overlooking rich biological and semantic information. 
(2) \textbf{Challenges in evaluation standardization}: Current studies face diverse real-world objectives (e.g., binary, multiclass, multilabel tasks) and research contexts. Therefore, they often adopt inconsistent experimental scenarios, varied evaluation metrics, and different baseline selections, making fair and standardized evaluation difficult. Existing benchmarks, such as DDI-Ben~\cite{ddi-ben}, evaluate generalization but overlook fundamental issues in data (e.g., relying on only 2 small datasets). Moreover, they include few algorithms and limited evaluation scenarios.

  \begin{figure}[t]
   \centering
   \includegraphics[width=0.9\linewidth]{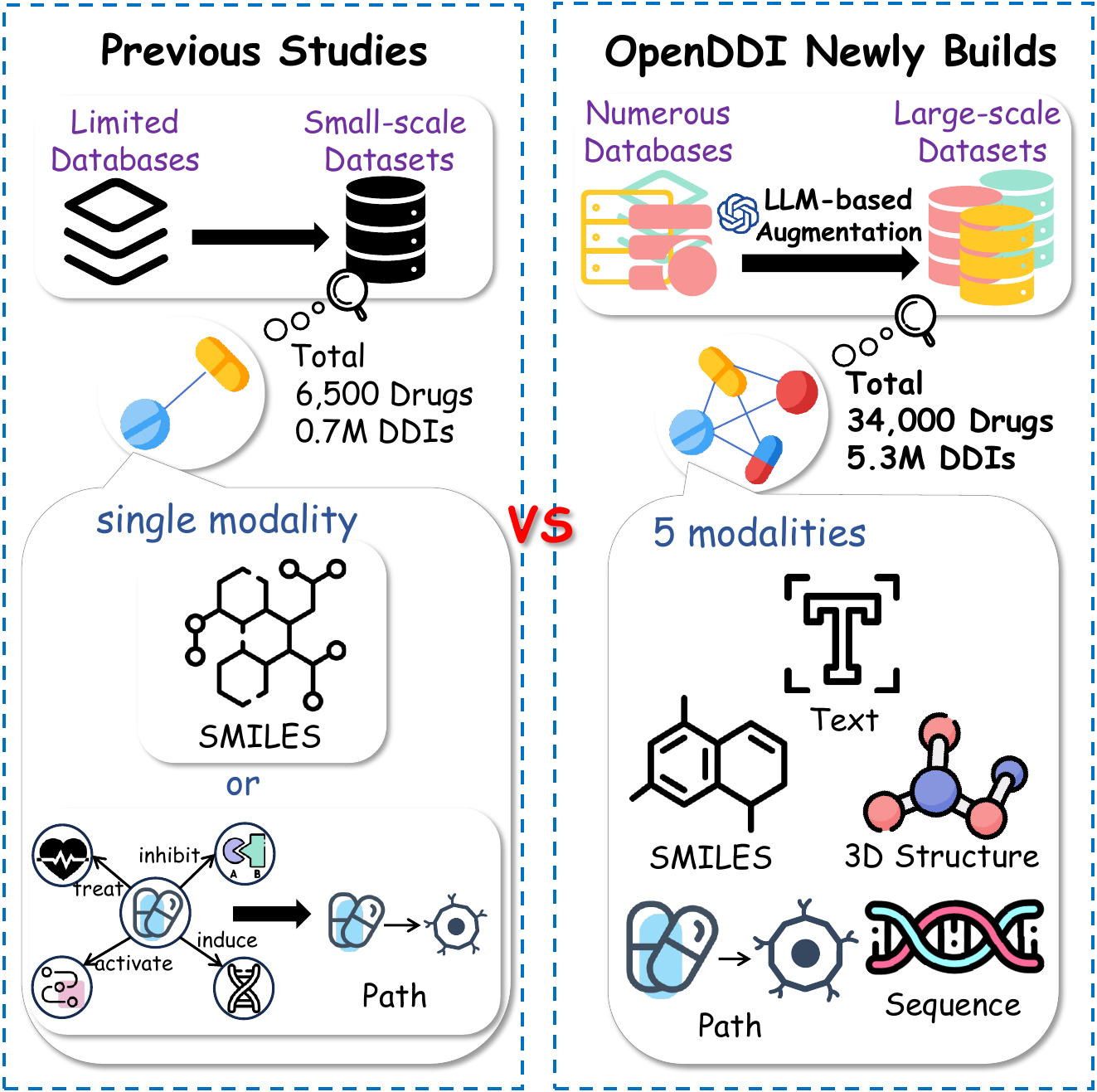}
   \vspace{-0.2cm}
   \caption{Comparison of previous and our newly-built datasets.}
   \vspace{-0.65cm}
   \label{data_cmp}
  \end{figure}

To address these limitations, we propose OpenDDI, a comprehensive benchmark for DDI prediction.
A systematic comparison with DDI-Ben is provided in Table~\ref{tab:DDI-benchmark-comparison}, which highlights advantages of OpenDDI in data coverage, algorithm breadth, task variety, and evaluation scenarios.
Following the comparison, we describe OpenDDI from two perspectives.
(1) From the \textbf{data} perspective, OpenDDI integrates representative DDI datasets and diverse drug representations.
Specifically, OpenDDI unifies 6 widely used DDI datasets and 2 existing forms of drug representation. Notably, OpenDDI additionally contributing 3 new large-scale datasets. New datasets are constructed through systematic curation and LLM-based augmentation. They contain over 5.3 million DDIs across 34000 drugs, significantly expanding the scale and diversity of available resources.
In addition to new datasets, OpenDDI newly builds multimodal drug representation.
The representation is constructed by integrating heterogeneous information about drugs, categorized into 5 modalities: SMILES, path, 3D conformations, amino acid sequences, and textual descriptions. Such integration enables richer understanding of drug properties and interactions.
Together, these datasets and representations provide diverse data construction options for DDI prediction.
They offer several advantages as demonstrated in Figure~\ref{data_cmp}.

(2) From the \textbf{evaluation} perspective, OpenDDI incorporates 20 SOTA baselines and 3 representative link-level downstream tasks. It further establishes standardized evaluation protocols and metrics for fair assessment across scenarios involving predictive performance, model robustness, and computational efficiency.
Based on OpenDDI, we provide a comprehensive evaluation of existing DDI prediction methods and datasets, drawing some valuable insights.
For \textbf{data quality}, we evaluate existing datasets from two perspectives: the scale and coverage of datasets, as well as the performance of various models trained on them. Through comparative analyses, we demonstrate the superiority of our curated datasets and further provide an in-depth investigation of the multimodal drug representations.
For \textbf{effectiveness}, we compare the performance of models across different datasets, comprehensively assessing the predictive capability of each model.
For \textbf{generalization}, we simulate realistic scenarios involving unseen drugs and evaluate the performance of models on them, thereby assessing their ability to predict beyond the training distribution.
For \textbf{efficiency}, we analyze model scalability with respect to both time and memory consumption. By examining empirical complexities, we highlight the computational efficiency and scalability of the models.
For \textbf{robustness}, we focus on challenges arising from data noise and sparsity, assessing how well existing approaches maintain performance when exposed to different types and degrees of data perturbation.

\textbf{Contributions}.
(1) \textit{\underline{Comprehensive Benchmark}}: OpenDDI unifies 6 existing datasets and 2 existing forms of drug representation, while additionally contributing 3 new LLM-augmented datasets and a new multimodal drug representation covering 5 biomedical modalities, broadening the current scope of DDI studies. Furthermore, OpenDDI integrates 20 SOTA algorithms and 3 downstream tasks for evaluations.
(2) \textit{\underline{Analytical Insights}}: Through extensive experimentation, we identified 11 pivotal insights from five analytical perspectives, which collectively deepen our understanding of DDI prediction and inspire methodological advancements.
(3) \textit{\underline{Open-Source Library and Detailed Repository}}: OpenDDI is designed as an open-access data and code platform, allowing users to easily evaluate their algorithms, while promoting innovation using the curated resources.

 \begin{figure*}[h]
  \centering
  \includegraphics[width=0.93\linewidth]{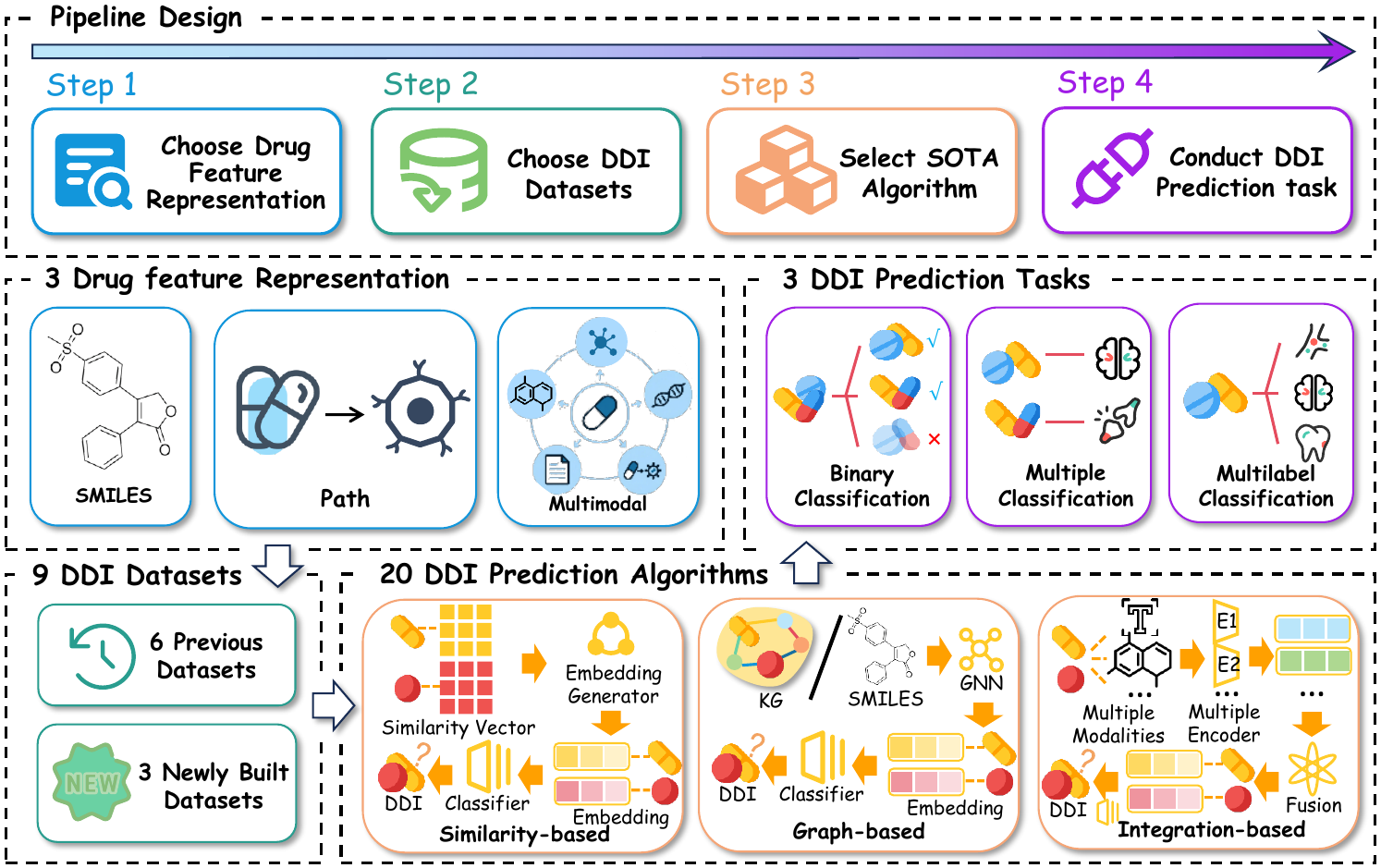}
  \vspace{-0.2cm}
  \caption{An overview of the OpenDDI Framework.}
  \vspace{-0.55cm}
  \label{framework}
 \end{figure*}

\section{Definitions and Background}

We formalize DDI prediction problem as follows.

\textbf{DDI Graph Formulation.} We model the DDI prediction task using an undirected attributed multi-relational graph $\mathcal{G} = (\mathcal{V}, \mathcal{E}, \mathbf{X})$. The node set $\mathcal{V} = \{v_1, \dots, v_n\}$ comprises $n$ distinct drugs. 
The edge set $\mathcal{E}$ characterizes interactions, indicating that drugs $v_i$ and $v_j$ exhibit interaction label(s) $y_{ij} \in \mathcal{Y}$, where $\mathcal{Y}$ denotes the label set.
Furthermore, $\mathbf{X} \in \mathbb{R}^{n \times d}$ serves as the node feature matrix, with the $i$-th row $\mathbf{x}_i \in \mathbb{R}^d$ encoding the feature representation of drug $v_i$.

\textbf{Drug Representation.} 
Each drug $v \in \mathcal{V}$ is associated with a feature representation. 
Former DDI studies typically rely on a single modality, such as (i) chemical structure encodings (e.g., SMILES) or (ii) path embeddings derived from drug–entity relations. In contrast, OpenDDI introduces a unified \emph{multimodal} representation by integrating heterogeneous evidence, including SMILES, path, 3D conformations, protein sequences, and textual descriptions. 
Formally, each drug is mapped to a feature tuple
\[
    \Phi(v) = \big(\mathbf{x}_v^{(1)}, \dots, \mathbf{x}_v^{(M)}\big),
\]
where $M$ is the number of modalities and $\mathbf{x}_v^{(m)} \in \mathbb{R}^{d_m}$ denotes the $m$-th modality-specific embedding. 
For downstream prediction, we further construct a fused feature vector 
\[
    \mathbf{x}_v = g\!\left(\mathbf{x}_v^{(1)}, \dots, \mathbf{x}_v^{(M)}\right) \in \mathbb{R}^d,
\]
where $g(\cdot)$ is a modality-aggregation function.  
The resulting fused feature vectors $\{\mathbf{x}_v\}_{v \in \mathcal{V}}$ constitute the node feature matrix $\mathbf{X} \in \mathbb{R}^{n \times d}$ defined in the graph formulation.

\textbf{Prediction Tasks.} 
The goal is to learn a predictor $f(v_i, v_j; \mathcal{G}) \mapsto \hat{y}_{ij} \in \mathcal{Y}$ using graph structure and node features. We consider three standard instantiations.
\textit{Binary interaction prediction}: $\mathcal{Y} = \{0,1\}$. $y_{ij}$ is a binary variable where $y_{ij} = 1$ denotes the presence of a DDI between drug $v_i$ and drug $v_j$, and $y_{ij} = 0$ denotes its absence.
\textit{Multiclass interaction classification}: $\mathcal{Y} = \{1,\dots,C\}$, where  
$y_{ij}=c$ specifies that the pair interacts through the $c$-th predefined interaction category.
\textit{Multilabel interaction classification}: $\mathcal{Y} = \{0,1\}^K$, where  $y_{ij} = (y_{ij}^{(1)}, \dots, y_{ij}^{(K)})$ is a multi-hot vector. For each mechanism index $k$, $y_{ij}^{(k)} = 1$ indicates the presence of the $k$-th interaction mechanism, while $y_{ij}^{(k)} = 0$ indicates that this mechanism does not occur.

\section{Benchmark Design}
\label{sec:benchmark_design}
\subsection{Algorithm Framework}

\textbf{DDI prediction Taxonomy.}  Existing DDI prediction methods can be systematically categorized into three paradigms according to their core modeling strategies: similarity-based, graph-based, and integration-based approaches. This comprehensive coverage reflects the major methodological directions in the field, as illustrated in Figure~\ref{framework}. 

\textbf{DDI prediction Algorithms.}  For DDI prediction algorithms, our framework incorporates 20 algorithms that are carefully implemented according to their original repositories or thoroughly interpreted from the corresponding papers. A full description of these algorithms is provided in Appendix~\ref{Baseline det}. In addition, we design a unified operational interface that integrates all models into a single API, enabling seamless invocation, systematic evaluation, and future extensibility. Through the efforts within OpenDDI, we offer a powerful platform that supports rigorous and trustworthy benchmarking for researchers.

\subsection{Dataset Curation}

\textbf{Data Components.}  
To comprehensively evaluate DDI prediction methods, we compile 9 datasets along with 3 types of drug feature representations. Specifically, the dataset includes 6 previously used and 3 newly curated large-scale datasets. For drug representations, we include 3 categories, namely SMILES, path, and a newly developed multimodal representation that spans five distinct types of information. Detailed descriptions are provided in Appendix~\ref{Data det}.

\textbf{Datasets newly built in OpenDDI.}  
In this section, we introduce the newly built DDI datasets and representation. We curates drug data from 12 authoritative databases (e.g., DrugBank~\cite{DrugBank2017}, ChEMBL~\cite{ChEMBLTD2018}, BindingDB~\cite{bindingdb2024}, KEGG~\cite{KEGG}), forming the largest DDI datasets and the richest modality coverage for drug feature representation.
Specially, (1) \textbf{For DDI datasets}, OpenDDI substantially enlarges dataset scale and coverage, containing 3 task-specific datasets with more than 5.3 million interactions across 34,000 drugs. Raw interactions and metadata are systematically integrated, and then annotated into binary, multiclass, and multilabel formats~\cite{DDIReview} to support diverse tasks. LLMs are employed to supplement uncertain labels, ensuring their completeness. Compared with existing datasets such as ChCh-Miner~\cite{StructNet-DDI} and Ryus~\cite{deepddi2018}, OpenDDI achieves an order-of-magnitude improvement in both the number of instances and the diversity of tasks, establishing a richer foundation for research.
Besides, (2) \textbf{For drug feature representation}, OpenDDI achieves a comprehensive representation of drugs by integrating five complementary modalities~\cite{Multimodal1,Multimodal3}, each providing a unique perspective on drug properties and interactions. Text contributes the semantic pharmacological knowledge, SMILES captures the structural similarity as a primary driver of DDIs, protein sequences and path encode the biochemical information, 3D molecular structures reflect the spatial conformations critical for binding affinity. Collectively, these modalities complement each other across semantic, structural, biological, and relational dimensions, thereby enabling models to capture the underlying logic of drug interactions from micro-level structures to macro-level mechanisms.

\subsection{Evaluation Procedure}
\textbf{Pipeline Design}. In previous studies, differences in DDI datasets, drug feature representations, and downstream tasks posed significant challenges to achieve fair comparison. To bridge this gap, we systematically modified and extended existing DDI prediction methods, achieving standardized data input through unified code interfaces and ensuring comprehensive coverage of downstream tasks. This pipeline enables thorough evaluation of the methods.

\textbf{Data Quality}. For the datasets and drug representations included in OpenDDI, we assess their quality from two perspectives: the scale and coverage of each dataset; the model performance across different datasets and drug representations. Comparative results highlight the advantages of our curated data. In addition, we conduct a detailed ablation study to further analyze the contribution of different components within the multimodal drug representations.

\begin{table*}[ht]
\centering
\fontsize{7.5}{8.5}\selectfont    
\renewcommand{\arraystretch}{1.3}  
\setlength{\tabcolsep}{4pt}        
\caption{Performance comparison of methods on datasets with MM (multimodal), SMILES, and Path representations. The impr. shows the average relative performance improvement of MM over SMILES and Path, with positive values (marked with $\uparrow$).}
\label{tab:performance_mm_smiles_path_relative}
\begin{tabular}{l!{\vrule width 1pt}cccc!{\vrule width 0.8pt}cccc!{\vrule width 0.8pt}cccc}
\toprule[1pt]
\multirow{2}{*}{\textbf{\vspace{-5pt} \; Method}} &
\multicolumn{4}{c|}{\textbf{ODDI-M}} &
\multicolumn{4}{c|}{\textbf{Ryus}} &
\multicolumn{4}{c}{\textbf{ZS-DDIE}} \\
\cmidrule(l){2-5} \cmidrule(l){6-9} \cmidrule(l){10-13}
& SMILES & Path & MM & impr. (\%) &
SMILES & Path & MM & impr. (\%) &
SMILES & Path & MM & impr. (\%) \\
\midrule[1pt]
DeepDDI &
35.5 {\tiny ±1.9} &
29.0 {\tiny ±1.8} &
53.3 {\tiny ±1.6} &
\textbf{68.8~$\uparrow$} &
70.8 {\tiny ±1.3} &
64.0 {\tiny ±1.4} &
78.3 {\tiny ±1.0} &
\textbf{16.9~$\uparrow$} &
53.2 {\tiny ±1.7} &
47.9 {\tiny ±1.8} &
67.5 {\tiny ±1.5} &
\textbf{33.7~$\uparrow$} \\
CASTER &
52.7 {\tiny ±1.8} &
52.8 {\tiny ±1.7} &
77.2 {\tiny ±1.3} &
\textbf{46.4~$\uparrow$} &
75.4 {\tiny ±1.2} &
73.3 {\tiny ±1.3} &
84.4 {\tiny ±0.9} &
\textbf{13.4~$\uparrow$} &
65.1 {\tiny ±1.6} &
59.4 {\tiny ±1.8} &
66.8 {\tiny ±1.5} &
\textbf{8.9~$\uparrow$} \\
\midrule[1pt]
GoGNN &
71.1 {\tiny ±1.4} &
67.4 {\tiny ±1.5} &
74.7 {\tiny ±1.2} &
\textbf{8.3~$\uparrow$} &
84.0 {\tiny ±0.9} &
80.6 {\tiny ±1.0} &
88.4 {\tiny ±0.7} &
\textbf{7.6~$\uparrow$} &
74.5 {\tiny ±1.3} &
67.3 {\tiny ±1.5} &
78.0 {\tiny ±1.2} &
\textbf{11.0~$\uparrow$} \\
MRCGNN &
76.5 {\tiny ±1.3} &
74.3 {\tiny ±1.4} &
81.6 {\tiny ±1.1} &
\textbf{8.4~$\uparrow$} &
85.1 {\tiny ±0.9} &
83.3 {\tiny ±1.0} &
92.6 {\tiny ±0.6} &
\textbf{10.1~$\uparrow$} &
75.9 {\tiny ±1.3} &
72.9 {\tiny ±1.4} &
79.9 {\tiny ±1.1} &
\textbf{7.8~$\uparrow$} \\
LAGAT &
71.4 {\tiny ±1.5} &
69.2 {\tiny ±1.6} &
73.4 {\tiny ±1.4} &
\textbf{4.5~$\uparrow$} &
86.7 {\tiny ±1.0} &
85.3 {\tiny ±1.1} &
90.5 {\tiny ±0.9} &
\textbf{5.3~$\uparrow$} &
79.2 {\tiny ±1.4} &
75.2 {\tiny ±1.6} &
83.5 {\tiny ±1.2} &
\textbf{8.7~$\uparrow$} \\
DDKG &
84.9 {\tiny ±1.1} &
83.0 {\tiny ±1.3} &
86.1 {\tiny ±1.0} &
\textbf{2.6~$\uparrow$} &
88.9 {\tiny ±0.8} &
87.6 {\tiny ±0.9} &
90.5 {\tiny ±0.7} &
\textbf{2.6~$\uparrow$} &
80.7 {\tiny ±1.3} &
77.3 {\tiny ±1.5} &
81.2 {\tiny ±1.2} &
\textbf{3.0~$\uparrow$} \\
KGNN &
67.2 {\tiny ±1.7} &
64.7 {\tiny ±1.8} &
71.5 {\tiny ±1.5} &
\textbf{8.6~$\uparrow$} &
84.6 {\tiny ±1.1} &
84.0 {\tiny ±1.2} &
84.9 {\tiny ±1.1} &
\textbf{0.7~$\uparrow$} &
78.0 {\tiny ±1.4} &
75.7 {\tiny ±1.6} &
80.1 {\tiny ±1.3} &
\textbf{4.4~$\uparrow$} \\
\midrule[1pt]
DDIMDL &
44.2 {\tiny ±2.0} &
43.0 {\tiny ±1.9} &
56.8 {\tiny ±1.8} &
\textbf{30.7~$\uparrow$} &
71.1 {\tiny ±1.5} &
70.4 {\tiny ±1.5} &
79.4 {\tiny ±1.2} &
\textbf{12.1~$\uparrow$} &
58.1 {\tiny ±1.7} &
55.9 {\tiny ±1.8} &
74.0 {\tiny ±1.4} &
\textbf{30.4~$\uparrow$} \\
TIGER &
42.1 {\tiny ±2.1} &
37.5 {\tiny ±2.0} &
62.8 {\tiny ±1.6} &
\textbf{60.0~$\uparrow$} &
65.1 {\tiny ±1.5} &
63.6 {\tiny ±1.6} &
84.6 {\tiny ±1.0} &
\textbf{31.9~$\uparrow$} &
51.2 {\tiny ±1.9} &
48.8 {\tiny ±1.9} &
56.2 {\tiny ±1.8} &
\textbf{13.1~$\uparrow$} \\
MVA-DDI &
58.5 {\tiny ±1.7} &
56.3 {\tiny ±1.7} &
67.4 {\tiny ±1.4} &
\textbf{17.5~$\uparrow$} &
75.5 {\tiny ±1.3} &
74.3 {\tiny ±1.4} &
83.6 {\tiny ±1.1} &
\textbf{11.7~$\uparrow$} &
67.0 {\tiny ±1.6} &
67.1 {\tiny ±1.7} &
72.6 {\tiny ±1.5} &
\textbf{8.3~$\uparrow$} \\
MKGFENN &
86.5 {\tiny ±1.0} &
85.6 {\tiny ±1.1} &
87.0 {\tiny ±0.9} &
\textbf{1.1~$\uparrow$} &
89.7 {\tiny ±0.8} &
87.5 {\tiny ±0.9} &
90.3 {\tiny ±0.7} &
\textbf{1.9~$\uparrow$} &
81.1 {\tiny ±1.2} &
81.5 {\tiny ±1.2} &
82.8 {\tiny ±1.1} &
\textbf{1.8~$\uparrow$} \\
MUFFIN &
66.9 {\tiny ±1.5} &
77.2 {\tiny ±1.3} &
80.2 {\tiny ±1.2} &
\textbf{12.6~$\uparrow$} &
87.5 {\tiny ±0.9} &
87.7 {\tiny ±0.8} &
88.4 {\tiny ±0.9} &
\textbf{0.4~$\uparrow$} &
77.2 {\tiny ±1.2} &
76.2 {\tiny ±1.4} &
77.9 {\tiny ±1.3} &
\textbf{0.7~$\uparrow$} \\
\bottomrule[1pt]
\end{tabular}
\vspace{-4mm}
\end{table*}

\textbf{Effectiveness}. To evaluate the effectiveness of methods, we conduct multi-level comparisons across three downstream tasks—binary, multiclass and multilabel classification. We uses three widely adopted evaluation metrics (ACC, F1, and AUC-ROC) to measure the model’s predictive ability.

\textbf{Generalization}. To evaluate the generalization of methods, we consider realistic scenarios where trained models are evaluated on DDI test sets containing unseen drugs, in order to assess their ability to predict DDIs involving novel drugs.

\textbf{Efficiency}. In evaluating the efficiency of methods, we focus on both time and space complexity. Time complexity is assessed through empirical analyses of computational capability, while space complexity emphasizes memory efficiency and storage requirements. These complementary dimensions provide a holistic view of model scalability and enable us to determine whether a method is suitable for deployment in resource-constrained scenarios.

\textbf{Robustness}. To assess methods' ability to withstand real-world challenges, we examine the robustness in two scenarios: data sparsity and noise. Noisy data are employed to evaluate the capacity of a model to remain stable under perturbed DDI data, whereas sparse data serve to investigate how models respond to rare DDI events. Both challenges represent critical obstacles that must be addressed to ensure reliable DDI prediction in practical applications.

\begin{table}[ht]
\vspace{-0.1cm}
\caption{Comparative statistics of existing DDI datasets}
\label{dataset_stats}
\centering
\fontsize{9}{8.5}\selectfont    
\renewcommand{\arraystretch}{1.33}  
\setlength{\tabcolsep}{4pt}        
\begin{tabular}{@{}l l r r l@{}}
\toprule[1.5pt]
\multicolumn{1}{c}{\textbf{Dataset}} & 
\multicolumn{1}{c}{\textbf{Task Type}} & 
\multicolumn{1}{c}{\textbf{Drugs}} & 
\multicolumn{1}{c}{\textbf{DDIs}} & 
\multicolumn{1}{c}{\textbf{Label Type}} \\
\midrule[1pt]
\multirow{3}{*}{OpenDDI} 
& Binary & 14,090 & 2,547,620 & Existence of DDI \\
& Multiclass & 14,086 & 2,534,350 & 167 DDI types \\
& Multilabel & 6,704 & 211,951 & 200 DDI types \\
\midrule[0.8pt]
ZhangDDI & Binary & 548 & 48,548 & Existence of DDI \\
ChCh-Miner & Binary & 1,514 & 48,514 & Existence of DDI \\
Ryus & Multiclass & 1,700 & 191,570 & 86 DDI types \\
Dengs & Multiclass & 570 & 37,264 & 65 DDI types \\
ZS-DDIE & Multiclass & 2,004 & 394,118 & 174 DDI types \\
Twosides & Multilabel & 350 & 19,535 & 100 DDI types \\
\bottomrule[1.5pt]
\end{tabular}
\vspace{-0.4cm}
\label{tab:dataset_stats}
\end{table}

\section{Experiments and Analyses}

In this section, we present a suite of experiments that systematically evaluate the quality of data and the effectiveness, generalization, efficiency, robustness of methods. Guided by a set of targeted questions, we aim to: (1) reveal differences in how existing datasets support model training and evaluation; (2) investigate the multimodal representation proposed in OpenDDI in depth; (3) examine methods in terms of their performance across diverse tasks and ability to cope with real-world difficulties.

For \textbf{data quality}, \textbf{Q1}: What are the scale and coverage of existing datasets? \textbf{Q2}: Can these datasets effectively support model training and evaluation? \textbf{Q3}: What insights can be further uncovered from the multimodal drug representation? For \textbf{effectiveness}, \textbf{Q4}: How do existing DDI prediction methods perform across different DDI prediction tasks? For \textbf{generalization}, \textbf{Q5}: How do existing DDI prediction methods perform when encountering unseen drugs? For \textbf{efficiency}, \textbf{Q6}: How do these methods perform regarding space and time consumption in practical scenarios? For \textbf{robustness}, \textbf{Q7}: How do DDI prediction methods handle real-world challenges such as data noise and data sparsity? 

\subsection{Datasets Comparison}
To address \textbf{Q1}, we conducted detailed curation and statistical analysis of existing DDI datasets.

From Table~\ref{tab:dataset_stats}, we derive several key observations. (1) The DDI data compiled in previous datasets involves no more than 2,100 drugs, whereas OpenDDI encompasses DDI data covering 14,090 drugs, substantially exceeding previous datasets in terms of drug coverage. (2) Previous DDI datasets typically contain no more than 0.4 million interactions, with most comprising around 80,000, whereas OpenDDI organizes over 2.5 million interactions, representing a substantial increase in scale. (3) Previous multiclass DDI datasets generally include fewer than 100 labels, although ZS-DDIE includes a larger variety of DDI types, it contains a substantial number of sparse types (with frequencies below 10). In contrast, OpenDDI provides 167 well-represented DDI labels with far fewer sparse types, ensuring that each label is supported by more than 50 records, thereby enhancing the statistical reliability and categorical balance of the dataset. (4) Previous DDI datasets are usually designed for a single downstream task, whereas OpenDDI includes datasets for all major downstream tasks—binary classification, multiclass classification, and multilabel classification—thus achieving the broadest task coverage.

Through the exploration of \textbf{Q1}, we draw two key conclusions. \textbf{C1}: \textit{OpenDDI substantially surpasses previous datasets in scale, with a far greater number of drugs and a much larger volume of DDI records.} \textbf{C2}: \textit{OpenDDI covers significantly more DDI types with reliable records, and offers broader applicability, as it provides datasets for all major DDI prediction tasks, thereby meeting a wider range of research and practical needs.}~\cite{twosides}

\begin{figure*}[ht]
    \centering
    \includegraphics[width=0.9\textwidth]
    {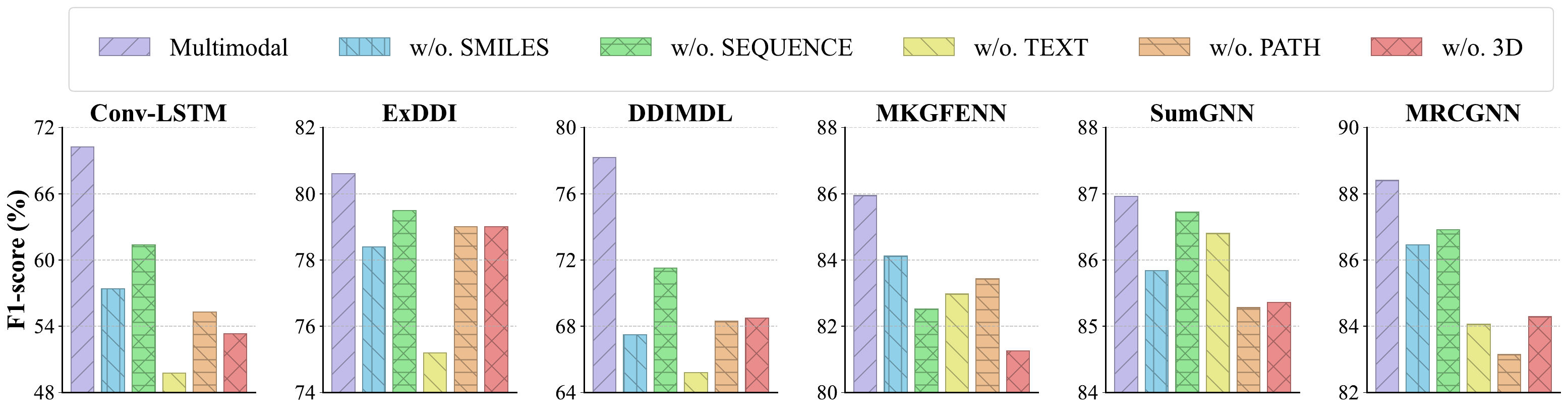}  
        \vspace{-0.3cm}
    \caption{F1-score(\%) comparison for DDI multiclass-prediction task after dissolve a modality.}
        \vspace{-0.3cm}
    \label{dissolve modality}
\end{figure*}

\begin{figure*}[ht]
    \centering
    \includegraphics[width=0.9\textwidth]
    {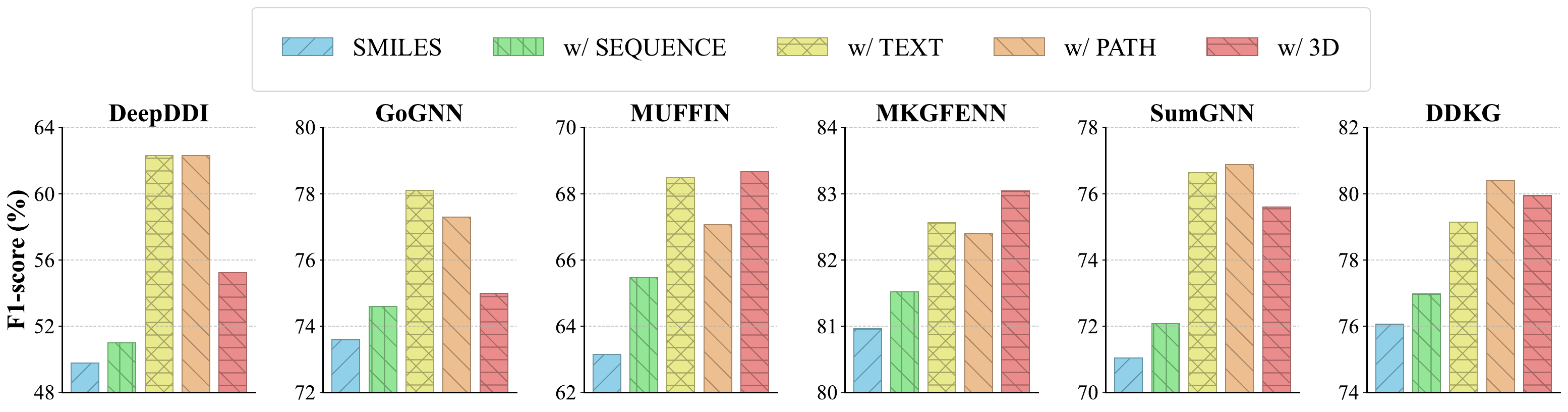}  
        \vspace{-0.3cm}
    \caption{F1-score(\%) comparison for DDI multiclass-prediction task with SMILES and another modality.}
    \vspace{-0.5cm}
    \label{SMILES and modality}
\end{figure*}

\vspace{-0.1cm}
\subsection{Datasets and Drug Representations Evaluation}
To address \textbf{Q2}, we evaluate the model performance across diverse DDI datasets and drug representations, as shown in Table~\ref{tab:performance_mm_smiles_path_relative}. From the perspective of DDI datasets, when facing more challenging multiclass prediction tasks, the performance of models trained and evaluated on OpenDDI datasets drops significantly, and the differences between models become more pronounced. This indicates that OpenDDI’s multiclass datasets are more challenging and better at differentiating the strengths and weaknesses of different methods. From the perspective of drug representations, methods that utilize multimodal drug features consistently outperform those based on other types of drug features, with the advantage being particularly significant in multiclass DDI datasets. This highlights the superiority of modeling drugs with multimodal features. 

Through the exploration of \textbf{Q2}, we can derive several key conclusions.
\textbf{C3}: \textit{For the more challenging DDI multiclass prediction tasks, existing methods perform poorly when facing difficult datasets, highlighting the need for further advancement in DDI prediction approaches}.
\textbf{C4}: \textit{Drug multimodal representations demonstrate clear advantages in drug modeling, underscoring the necessity of constructing multimodal representations}~\cite{mkgfenn2024}.

\begin{figure*}[ht]
    \centering
    \includegraphics[width=0.90\textwidth]
    {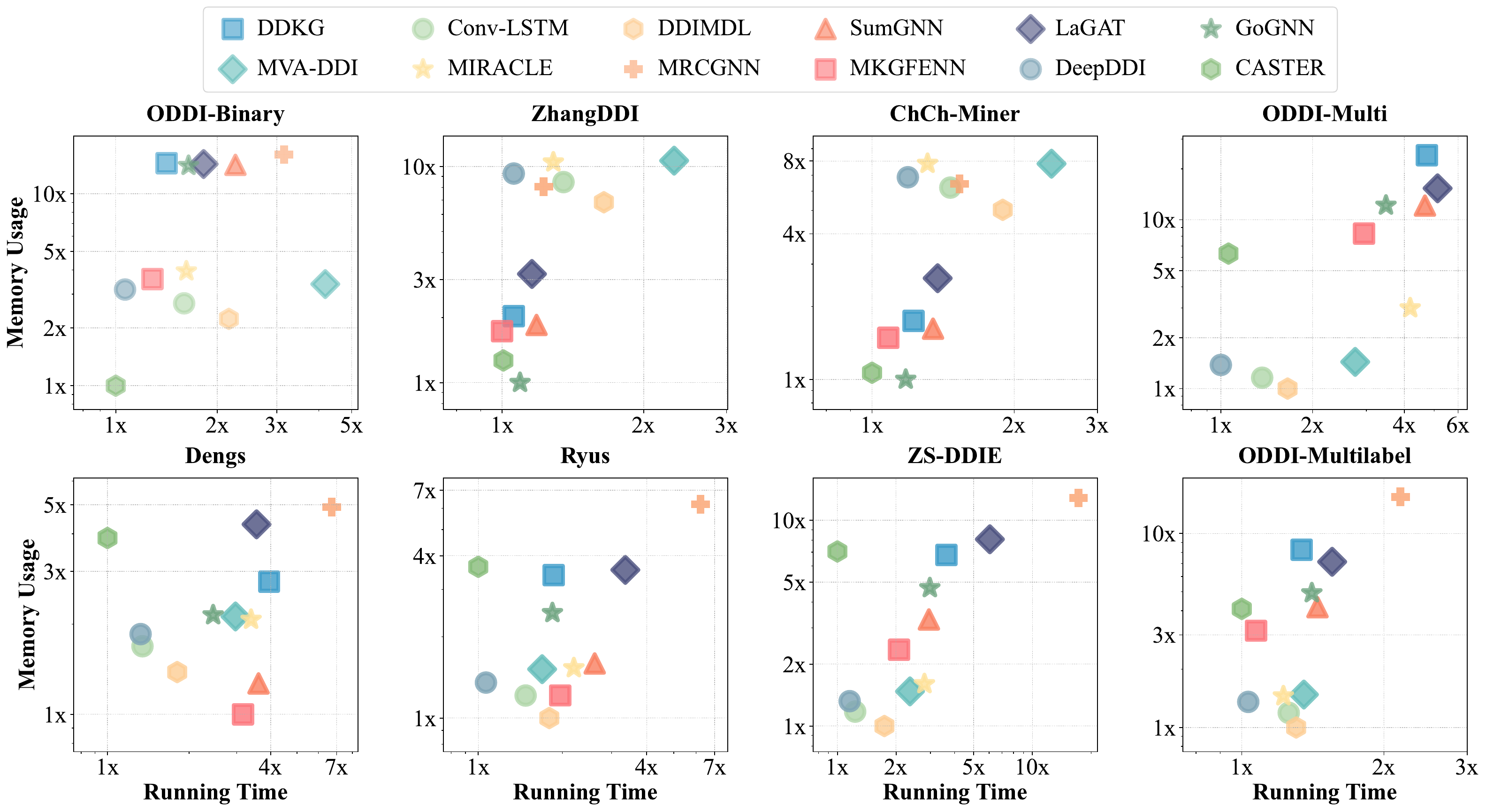}  
        \vspace{-0.25cm}
    \caption{Efficiency Performance on Various Datasets.}
        \vspace{-0.45cm}
    \label{Efficiency}
\end{figure*}

\subsection{Multimodal Research}
To address \textbf{Q3}, we conduct an in-depth investigation of the multimodal representations from two perspectives: (1) the necessity of each modality in drug modeling, and (2) the capability of other modalities to supplement perspectives absent in the SMILES modality. 

For the necessity of each modality in drug modeling, we obtain several observations from Figure~\ref{dissolve modality}:
(i) Compared with the complete multimodal presentations, the absence of any individual modality negatively impacts F1 scores in the DDI prediction task.
(ii) In particular, when SMILES, text, 3D, or path modality is removed, most models exhibit a marked drop in task performance.
(iii) By contrast, the absence of sequence modality causes a relatively smaller decline in performance for roughly half of the models.

For the capability of other modalities to supplement perspectives absent in the SMILES modality, we also derive several observations from Figure~\ref{SMILES and modality}:
(i) The integration of SMILES with other modalities consistently enhances F1 scores, demonstrating improved performance in the DDI prediction task.
(ii) Incorporating text, 3D, or path modalities in addition to SMILES leads to substantial improvements across most models.
(iii) However, compared with other modalities, the sequence modality contributes less pronounced benefits when integrated with SMILES for drug modeling learning.

Through our investigation of \textbf{Q3}, we arrive at three main conclusions. \textbf{C5}: \textit{All five modalities provide complementary perspectives for drug modeling, collectively enhancing the ability of DDI prediction models to capture drug characteristics and improve task performance}. \textbf{C6}: \textit{Among them, text, 3D, path, and SMILES prove most influential, with text, 3D, and path particularly reinforcing SMILES by supplying additional information and perspectives}. \textbf{C7}:\textit{ By contrast, the sequence modality contributes less, likely because drug–drug interactions are more strongly determined by structural properties and shared protein interactions, which tend to be better captured by other modalities}~\cite{ddkg2022, Multimodal3}.

\subsection{Model Performance Comparison}

To address \textbf{Q4}, we focus on Table~\ref{tab:performance_mm_smiles_path_relative}, Table~\ref{tab:performance_oddi_b_chch_simple} and Table~\ref{tab:performance_oddi_ml_twosides}. In the binary classification task, Graph-based and Integration-based methods generally perform well, with models such as DDKG and Muffin standing out. In the multiclass tasks, Graph-based methods excel by leveraging graph structures to enhance the understanding of drugs and DDIs, highlighting the importance of graph-aware modeling. Conversely, similarity-based methods exhibit limited predictive capacity. In the multilabel classification task, the differences among the three types of methods are relatively small.

\begin{table}[ht]
\centering
\fontsize{8.5}{11}\selectfont
\renewcommand{\arraystretch}{1.2}
\vspace{-0.2cm}
\caption{Generalization performance of methods when training set drugs are missing (w/: with training drugs; w/o.: without training drugs). The best results in each column are in \textbf{bold}.}
\label{tab:generalization_missing_drugs}
\begin{tabular}{l!{\vrule width 1pt}cc!{\vrule width 0.8pt}cc}
\toprule
\multirow{2}{*}{\textbf{\vspace{-5pt} \; Method}} &
\multicolumn{2}{c|}{\textbf{Ryus}} &
\multicolumn{2}{c}{\textbf{ZS-DDIE}} \\
\cmidrule(l){2-3} \cmidrule(l){4-5}
& w/ & w/o. & w/ & w/o. \\
\midrule
DeepDDI &
77.9 {\tiny ±1.2} & 58.5 {\tiny ±2.8} & 72.0 {\tiny ±1.8} & 25.9 {\tiny ±3.2} \\
CASTER &
87.7 {\tiny ±1.0} & \textbf{68.5} {\tiny ±2.4} & 77.8 {\tiny ±1.5} & \textbf{55.8} {\tiny ±2.9} \\
DSNDDI &
86.4 {\tiny ±1.1} & 56.3 {\tiny ±2.7} & 75.1 {\tiny ±1.6} & 40.1 {\tiny ±3.1} \\
MRCGNN &
88.5 {\tiny ±0.9} & 55.7 {\tiny ±2.6} & 80.4 {\tiny ±1.4} & 35.5 {\tiny ±3.0} \\
MKGFENN &
\textbf{89.6} {\tiny ±0.8} & 41.9 {\tiny ±3.1} & \textbf{82.1} {\tiny ±1.3} & 37.8 {\tiny ±3.3} \\
MVA-DDI &
86.6 {\tiny ±1.0} & 61.6 {\tiny ±2.5} & 81.9 {\tiny ±1.2} & 47.4 {\tiny ±2.8} \\
\bottomrule
\end{tabular}
\vspace{-0.3cm}
\end{table}

Through the exploration of \textbf{Q4}, we draw the conclusion \textbf{C8}:
 \textit{In DDI prediction tasks, Graph-based and Integration-based methods hold certain advantages. In contrast, similarity-based methods often yield sub-optimal performance, failing to capture high-order relational patterns}~\cite{mva2023}.

\begin{figure*}[ht]
    \centering
    \includegraphics[width=0.94\textwidth]
    {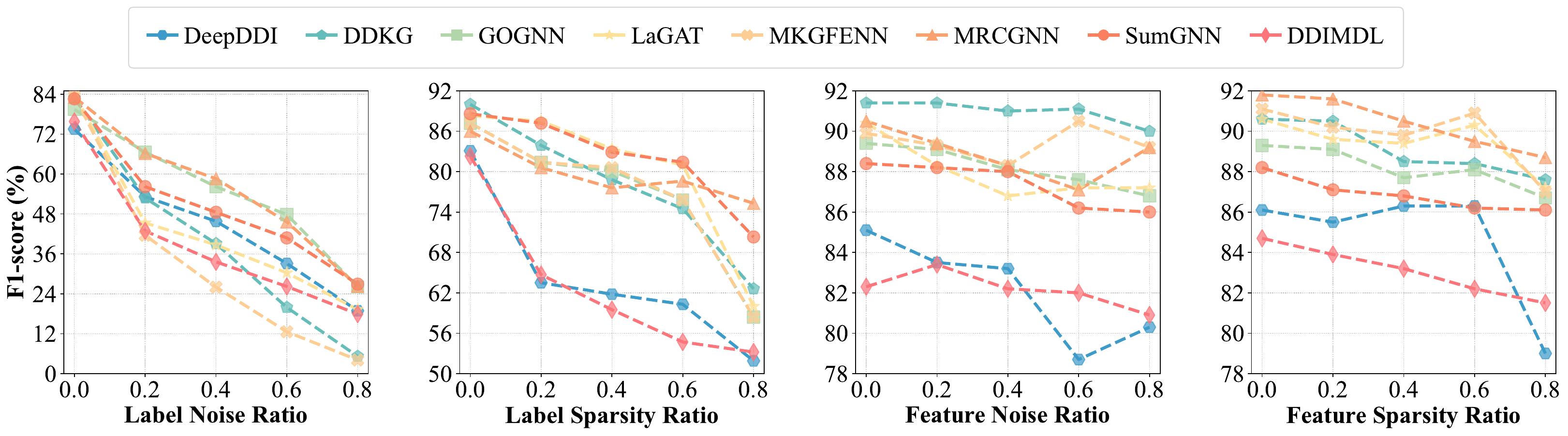}  
        \vspace{-0.35cm}
    \caption{Performance under Different Noise and Sparsity Ratios at Label and Feature Levels.}
    \label{robustness}
         \vspace{-0.6cm}
\end{figure*}

\subsection{Generalization Analyses}

To address \textbf{Q5}, we withhold 30\% drugs from the training set to serve as unseen drugs during testing, the detailed experimental results are summarized in Table~\ref{tab:generalization_missing_drugs}. Observations indicate a universal performance drop across all methods when they encounter previously unseen drugs. Notably, DeepDDI, MRCGNN, and MKGFENN exhibit a substantial decline on the ZS-DDIE dataset, with their performance metrics dropping by more than twofold. Conversely, CASTER maintains a degree of stability across both datasets, demonstrating superior generalization capabilities.

Through the exploration of \textbf{Q5}, we draw the conclusion \textbf{C9}: \textit{Existing methods exhibit insufficient generalization capabilities. To address the challenges of DDI prediction for unseen drugs, it is imperative for methods to bolster their generalization}~\cite{caster2020}.

\subsection{Efficiency Analyses}
To address \textbf{Q6}, we evaluated the efficiency of DDI prediction methods across a series of datasets from perspectives of memory and time, thereby analyzing their scalability.

As shown in Figure~\ref{Efficiency}, we can observe several notable trends. Graph-based methods, due to their reliance on additional operations over graph structures, generally incur higher memory in most experiments. Similarity and Integration-based methods, in contrast, exhibit lower memory usage on smaller datasets such as Dengs. However, when scaled to larger datasets like OpenDDI-Multi, these methods still demonstrate instability in memory consumption. With respect to computational time, most models achieve satisfactory efficiency on smaller datasets. Yet, when confronted with large-scale datasets, certain models such as MKGFENN incur substantially higher time costs.

Based on the analysis of \textbf{Q6}, we derived conclusion \textbf{C10}:\textit{ Existing methods require further reductions in memory and time to enhance model scalability, thereby addressing more challenging resource constraints}~\cite{lagat2022}.

\subsection{Robustness Analyses}
To address \textbf{Q7}, we designed comprehensive experiments to simulate real-world challenges in DDI prediction task, including data noise and sparsity. These experiments evaluated the performance of methods under realistic conditions, ensuring a thorough assessment of their robustness.

The experimental results (Figure~\ref{robustness}) show that noise and sparsity scenarios negatively affect the performance of all DDI prediction methods. In the case of label noise, increasing levels of noise consistently lead to sharp and steady declines in the performance of all algorithms. Regarding label sparsity, MRCGNN and SumGNN manage to maintain performance to some extent, whereas most other methods eventually exhibit substantial degradation; notably, DeepDDI and DDIMDL already experience significant drops under relatively low levels of label sparsity. Feature noise has a milder impact on model performance, and algorithms such as MKGFENN even show slight improvements under certain proportions of feature noise. As for feature sparsity, its effect on most methods is relatively limited, with the exception of DeepDDI, whose performance deteriorates markedly under high feature sparsity.

Based on the analysis of \textbf{Q7}, we draw the conclusion \textbf{C11}: \textit{the robustness of existing DDI prediction methods remains insufficient, particularly in handling label-related issues. Therefore, enhancing the robustness of these methods is an urgent need that must be addressed}~\cite{gognn2020}.

\section{Conclusion and Future Directions}
In this paper, we first introduce DDI prediction, outlining its application scenarios and recent research progress. We then present OpenDDI, a comprehensive benchmark for DDI prediction research, which unifies 6 widely used DDI datasets and 2 existing forms of drug representation, while additionally contributing 3 new large-scale LLM-augmented datasets and a new multimodal drug representation covering 5 modalities.
Besides, it unifies 20 baselines across 3 downstream tasks, with standardized protocols for data quality, effectiveness, generalization, robustness, and efficiency. With the standardized evaluation by OpenDDI, we conduct an in-depth investigation of DDI prediction, summarizing 11 valuable conclusions. Building upon these conclusions, we further identify promising future directions.

\textbf{Advancing the data development} (\textit{C1}, \textit{C2}, \textit{C3}, and \textit{C4}). Most existing DDI datasets suffer from limited scale and narrow application scenarios, while current drug representations often contain coarse or single-dimensional information. In fact, the scale and diversity of datasets determine the scope and applicability of DDI prediction tasks, whereas the quality of representations directly affects the models’ understanding of DDIs. For the sustainable development of DDI prediction, it is essential to advance the data development.

\textbf{Enhancing the interpretability of DDIs} (\textit{C5}, \textit{C6}, and \textit{C7}). Understanding how DDIs occur is of great significance both for DDI prediction and for real-world applications. The introduction of multimodal information offers additional perspectives for interpretability, enabling deeper insights into the underlying causes of DDIs. Future work on DDI prediction can further advance interpretability, providing novel perspectives on DDI mechanisms.

\textbf{Improving the predictive capability of methods} (\textit{C8} and \textit{C9}). Existing methods demonstrate promising performance on some earlier datasets, yet they often fail to sustain strong results when confronted with more challenging datasets or unseen drugs. To achieve more comprehensive and outstanding DDI prediction, and to address increasingly difficult challenges in the future, it is necessary to further strengthen both the predictive capability and adaptability of models.

\textbf{Optimizing the efficiency of algorithms} (\textit{C10}). Current DDI prediction methods incur substantial time and memory costs when handling large-scale datasets. As the field continues to advance, future datasets will inevitably grow in scale, further intensifying the challenge of algorithmic efficiency. To support the continued development of DDI prediction, it is therefore imperative to optimize algorithm efficiency and enhance model scalability.

\textbf{Addressing real-world DDI data challenges} (\textit{C11}). In practical DDI prediction scenarios, data noise and sparsity are unavoidable obstacles. Current DDI prediction methods remain insufficiently adapted to these real-world challenges, particularly when handling noisy and sparse label. Future methods should place greater emphasis on coping with such challenges and enhancing robustness.

\section*{Impact Statement}

This work aims to advance DDI prediction research by introducing a comprehensive benchmark that integrates diverse datasets and multimodal drug representations. By promoting standardized and reproducible evaluation, our work may contribute to the development of more reliable DDI prediction models, which could ultimately support medication safety, pharmacovigilance, and drug discovery.

Potential risks include the possibility that models evaluated on OpenDDI could be mistakenly treated as clinical decision tools without sufficient validation, or that dataset biases may lead to uneven performance. To mitigate these concerns, OpenDDI is intended strictly for research purposes, and we emphasize that DDI predictions should not replace expert judgment or clinical validation. We also encourage transparent reporting, careful dataset documentation, and responsible deployment of predictive models.

Overall, we believe this work supports responsible progress in AI-driven drug research while highlighting the importance of cautious and ethical use in healthcare contexts.


\bibliography{ref}
\bibliographystyle{icml2025}

\newpage
\appendix
\onecolumn

\lstset{
    language=Python,
    basicstyle=\ttfamily\small,
    keywordstyle=\bfseries,
    stringstyle=\itshape\color{gray!80!black},
    commentstyle=\color{gray},
    numbers=none,
    numberstyle=\tiny,
    showstringspaces=false,
    breaklines=true,
    frame=lines,      
    captionpos=t,
    escapeinside={(*@}{@*)}
}

\section{Baseline Algorithm Details}
\label{Baseline det}

\textbf{DeepDDI}~\cite{deepddi2018} is a classic deep learning-based benchmark model for drug interaction prediction. It takes chemical structure information (SMILES) and names of drug pairs as input, extracts features through a structural similarity profile (SSP), and employs a deep neural network for multi-label classification, outputting human-readable sentences that describe the interaction types. The architecture of this method is relatively straightforward, providing a reliable and commonly used performance reference point for evaluating more complex drug interaction prediction models.

\textbf{DDIMDL}~\cite{ddimdl2020} is a multimodal benchmark model designed to predict specific events of drug-drug interactions. It constructs individual deep neural network sub-models for different drug features (such as chemical substructures, targets, and enzymes), then fuses the outputs of these sub-models to learn cross-modal representations of drug pairs. This method demonstrates excellent performance on standard datasets. Compared to approaches that rely solely on single-source features (e.g., DeepDDI), it provides an important reference on how to effectively integrate multi-source data for fine-grained prediction.

\textbf{CASTER}~\cite{caster2020} is an advanced model for predicting drug interactions that emphasizes interpretability. It extracts functional substructures from drug chemical structures via sequence mining and pre-trains an autoencoder on large-scale unlabeled data to obtain generalized representations. Its core innovation lies in projecting drug pairs into a substructure space through dictionary learning, where the resulting sparse coefficients are used both for prediction and for quantifying the importance of each substructure. This achieves a unification of performance and chemical interpretability, establishing an important benchmark for such research.

\textbf{DDKG}~\cite{ddkg2022} is an attention-based knowledge graph representation learning model for predicting drug interactions. It synergistically leverages the chemical structures (SMILES) of drugs and knowledge graph triples, first learning initial embeddings via an encoder-decoder and then propagating and aggregating neighbor information along high-attention paths to derive global node representations. The model serves as a strong benchmark for fusing multi-source knowledge for relational inference.

\textbf{KGNN}~\cite{kgnn2020} is a knowledge graph‑based graph neural network framework for drug‑drug interaction prediction. Unlike methods that rely on chemical structures or single‑source features, KGNN learns high‑order topological representations of drugs and their related entities (e.g., targets, pathways) from knowledge graphs through multi‑hop neighborhood sampling and aggregation. The model achieves strong performance across multiple drug datasets, particularly excelling in modeling long‑range semantic relationships, and serves as an important benchmark for knowledge‑graph‑based drug reasoning.

\textbf{LaGAT}~\cite{lagat2022} is a graph attention network model for drug-drug interaction prediction. Its core innovation lies in a link-aware attention mechanism, which dynamically adjusts node attention pathways based on the currently predicted drug pair, thereby capturing the semantic diversity of the same drug in different interaction contexts. This method achieves excellent predictive performance and interpretability on knowledge graphs, providing an important technical reference for drug relation inference.

\textbf{ExDDI}~\cite{exddi2025} is a model series designed to generate natural language explanations for drug-drug interaction prediction. It extends the DDI task for the first time to a joint learning task of "prediction + explanation generation." The series leverages the molecule-text pretrained model MolT5, exploring various methods such as sequence-to-sequence generation, multi-task training, retrieval, and large language model in-context prompting. Experiments demonstrate that training with more detailed explanations not only enhances interpretability but also improves prediction performance, providing an important benchmark for explainable drug safety research.

\textbf{GOGNN}~\cite{gognn2020} is a graph-of-graphs neural network designed for drug-drug interaction (DDI) prediction. The model represents each entity as a local graph (e.g., molecular structure) and models inter-entity interactions as a global graph. It employs attention-based pooling on local graphs to extract key substructures and uses attention aggregation on the global graph to integrate interaction information. GoGNN demonstrates strong performance across multiple DDI prediction tasks, serving as a key benchmark method for graph-structured drug relationship modeling.

\textbf{MIRACLE}~\cite{miracle2021} is a multi-view graph contrastive learning model for drug-drug interaction prediction. It encodes drug molecular structures using a bond-aware message passing network (BAMPN), integrates drug interaction network information via a graph convolutional network (GCN), and employs contrastive learning to align representations from both molecular structure and interaction relationship views. The model demonstrates robust performance across datasets of varying scales and exhibits good scalability. As a representative multi-view learning approach, MIRACLE is often adopted as a benchmark to evaluate the ability of new methods in integrating structural and relational information.

\textbf{TIGER}~\cite{tiger2024} is a dual-channel drug-drug interaction prediction model based on a relation-aware graph Transformer. It models multi-semantic relations between nodes via a relation-aware self-attention mechanism and employs subgraph sampling strategies to handle large-scale knowledge graphs. The model integrates dual-channel information from both drug molecular graphs and biomedical knowledge graphs, effectively capturing long-range dependencies and high-order structures. It demonstrates strong adaptability and predictive performance across datasets of varying scales, serving as a benchmark for evaluating Transformer-based multi-relational graph learning methods.
 
\textbf{SumGNN}~\cite{sumgnn2021} is a graph representation learning framework based on knowledge graph summarization for multi-type drug-drug interaction prediction. It employs local subgraph extraction and layer-independent self-attention to generate sparse, interpretable reasoning pathways, and integrates multi-channel information (such as subgraph features and drug fingerprints) to enhance predictive performance—especially in low-resource scenarios. This model serves as an important benchmark for knowledge graph-enhanced DDI prediction.

\textbf{PHGLDDI}~\cite{phglddi2025} is a hierarchical graph neural network framework designed for drug-drug interaction (DDI) prediction. The model adopts a dual-level architecture: at the molecular level, it extracts structural features of drugs using a pre-trained Graph Convolutional Network (GCN); at the network level, it learns the topological relationships between drugs in the interaction network through a Graph Isomorphism Network (GIN). By integrating molecular structural information with network structural information, PHGL-DDI demonstrates strong generalization capability in inductive learning tasks (predicting new drugs), making it a representative method in graph neural network-based DDI prediction.

\textbf{MRCGNN}~\cite{mrcgnn2023} is a graph neural network model designed for predicting drug–drug interaction events. The model integrates structural information from drug molecular graphs with interaction information from DDI event graphs, employing a multi-relational contrastive learning approach and a dual-view negative augmentation strategy, which significantly enhances its ability to predict rare DDI events. MRCGNN not only demonstrates the advantages of jointly leveraging structural and interaction information, but also serves as an important example of using contrastive learning to address long-tailed distribution in graph data. It is regarded as a representative work in graph neural network-based DDI prediction.

\textbf{DSNDDI}~\cite{dsnddi} is a state-of-the-art graph neural network-based framework for drug-drug interaction prediction that emphasizes dual-view representation learning. It models drugs as molecular graphs and iteratively learns substructures from both intra-view (within a single drug) and inter-view (between drug pairs) perspectives using a series of DSN encoders, which integrate graph attention mechanisms for local atom-level updates and self-attention pooling for global drug representations. The DSN decoder employs a co-attention scoring function to predict interactions by integrating pairwise substructure embeddings, enabling high accuracy in both transductive and inductive settings without relying on additional domain knowledge.

\textbf{ConvLSTM}~\cite{convlstm2019} is a drug-drug interaction prediction model that combines knowledge graph embeddings with a convolutional long short-term memory (Conv-LSTM) network. This method learns drug representations from multi-source knowledge graphs, extracts local features through convolutional layers, and then integrates them via LSTM layers for final prediction. Conv-LSTM excels in integrating heterogeneous drug information and handling high-dimensional sparse data, making it an early representative work in knowledge graph-based DDI prediction.

\textbf{MUFFIN}~\cite{muffin2021} is a multi-scale feature fusion model for drug-drug interaction prediction. It integrates molecular structure features of drugs with semantic information from knowledge graphs, employing a dual-level fusion strategy (cross-level and scalar-level) to effectively combine multimodal information. MUFFIN supports binary, multi-class, and multi-label prediction tasks, demonstrating strong performance across multiple benchmark datasets. It is regarded as an early representative approach for fusing molecular graph and knowledge graph information in DDI prediction.

\textbf{MVA}~\cite{mva2023} is a multi-view attention network model designed for drug-drug interaction prediction. It employs Transformer and Graph Convolutional Networks to extract sequence and structural features from drugs, respectively, and utilizes contrastive learning-based self-supervised pretraining to enhance graph representation. The model further adaptively fuses the dual-view features through an attention mechanism, generating discriminative drug representations. This approach demonstrates excellent performance across multiple benchmark tasks and serves as a representative example of integrating self-supervised pretraining with attention-based fusion in multi-view representation learning.

\textbf{MKGFENN}~\cite{mkgfenn2024} is an end-to-end neural network model for DDI prediction that fuses multimodal knowledge graphs. It integrates four types of knowledge graphs—drug–chemical entities, drug–substructures, drug–drug interactions, and molecular structures—and employs a four-channel graph neural network to extract features, which are then fused via a multilayer perceptron in an end-to-end manner. The model demonstrates strong performance across a variety of DDI prediction tasks, serving as a representative approach for combining multi-source knowledge through end-to-end joint learning.

\textbf{MMDGDTI}~\cite{mmdgdti2025} is a DTI prediction model that integrates multimodal feature fusion and domain generalization. It employs a large language model-based textual feature extractor and a hybrid graph neural network-based structural feature extractor to extract deep semantic and topological features of drugs, while enhancing generalization capability through domain adversarial training and contrastive learning. The model demonstrates strong performance in both single-domain and cross-domain tasks, serving as a multimodal prediction method with notable domain adaptability.

\textbf{ZeroDDI}~\cite{zeroddi2024} is a method designed for zero-shot drug-drug interaction event prediction. It learns discriminative features by integrating biologically enhanced semantic representations of DDI events with molecular substructure-guided attention, and introduces a dual-modal uniform alignment strategy to mitigate class imbalance. The method demonstrates strong performance in zero-shot prediction scenarios.

\section{Dataset details}
\label{Data det}

The three core datasets in the OpenDDI benchmark were constructed by integrating and harmonizing data from 12 authoritative source databases, followed by a significant LLM-assisted augmentation process. This yielded a unified, multimodal graph encompassing 14,324 drugs and over 2.54 million interaction edges, which substantially expanded the coverage (e.g., drugs increased from 4,467 to 14,090 and DDI relations from 1.4M to 2.55M in the binary task). To elucidate the data composition, the following section details these 12 source databases, outlining their respective roles in providing multimodal node features and DDI edges, along with their inherent characteristics. Subsequently, five external DDI datasets used for comparative evaluation are introduced. A comprehensive quantitative comparison between the OpenDDI datasets and these external benchmarks is provided in Table \ref{tab:dataset_stats}.

\subsection{Source Databases for Graph Construction}
\textbf{BindingDB}~\cite{bindingdb2006, bindingdb2024} is a public database of measured binding affinities, focusing on protein-target interactions with drug-like molecules. It provided detailed Kd, Ki, and IC50 values, which were instrumental in building the quantitative mechanism protein interaction modality, informing the model about the strength and specificity of drug-target binding.

\textbf{The BIOSNAP Drug-Drug Interaction (DDI) dataset}~\cite{biosnapnets} is a network of known interactions between FDA-approved drugs, containing 1,514 drugs and 48,514 interactions extracted from drug labels and scientific publications. It provides validated DDI edges for graph construction and network statistics, serving as a valuable resource for studying drug interactions.

\textbf{ChEMBL}~\cite{ChEMBLTD2018} is a large, manually curated bioactivity database containing over 1.8 million drug-like compounds and more than 15 million bioactivity records (as of Release 24, 2018). It serves as a primary source for molecular structures (e.g., SMILES strings) and bioactivity data, which are essential for constructing the 2D molecular graph modality.

\textbf{Davis}~\cite{Davis} dataset provides quantitative binding affinity measurements (Kd values) between kinases and their inhibitors. These continuous scores were used to weight and refine the mechanism protein interaction modality, adding a layer of quantitative confidence to drug-target links and offering a granular perspective beyond binary relationships.

\textbf{Drug Repurposing Knowledge Graph (DRKG)}~\cite{drkg2020} is a large, heterogeneous biological knowledge graph integrating genes, compounds, and diseases across 107 relationship types. It was mined for relationships between drugs and biological entities, contributing to the textual description and mechanism protein interaction modalities with rich, interlinked contextual information.

\textbf{DrugBank}~\cite{DrugBank2017} provides comprehensive information on drug–drug interactions, including interaction types, affected targets, and metabolic enzymes. It served as the central pillar for data integration, providing a primary drug ID system, core DDI edges, textual descriptions, molecular structures (SMILES, SDF), and mechanism protein interactions for thousands of approved and investigational drugs.

\textbf{KEGG}~\cite{KEGG} contains known adverse drug–drug interactions annotated as Contraindications (CI) and Precautions (P), providing systematic annotations of structural information, drug targets, and molecular mechanisms (e.g., CYP enzyme involvement) to support both 2D/3D structure‑based and textual description‑based modalities

\textbf{MEDLINE} is a premier bibliographic database from the U.S. National Library of Medicine. Its vast corpus of life sciences literature was the foundational source for the textual description modality, used to generate structured, informative textual profiles for each drug.

\textbf{OFFSIDES} is a resource containing statistically associated adverse drug reactions derived from FDA Adverse Event Reporting System (FAERS) reports, after filtering out known side effects already listed on FDA-approved labels. It was used to enrich the textual description modality by adding off-label safety information to drug profiles.

\textbf{Pharmacogenomics Knowledge Base (PharmGKB)}~\cite{PharmGKB} curates knowledge on how genetic variation impacts drug response. It provides annotated drug–gene relationships and pharmacogenomic pathway information, which was used to enrich the textual description modality and support the analysis of drug mechanism networks.

\textbf{PubChem}~\cite{PubChem2016} is a large-scale public repository of chemical substances and their biological activities. It contains over 60 million unique chemical structures (CIDs), each associated with multiple representations including SMILES, InChI, 2D/3D structures, and extensive bioactivity annotations. It is a major source for standardizing and retrieving molecular structure information, ensuring comprehensive coverage in the 2D molecular graph modality.

\textbf{TWOSIDES dataset}~\cite{twosides} is a resource for drug–drug–effect relationships, focusing on polypharmacy side effects. It serves as a core source for the multi-label classification dataset and providing multi-hot labels and DDI edges.

\subsection{External Datasets for Benchmarking}
\textbf{ChCh-Miner dataset}~\cite{StructNet-DDI} is a medium‑scale DDI benchmark comprising 1,514 drugs and 48,514 DDI links. It is widely adopted for evaluating the generalization ability of DDI prediction models across distinct drug networks.

\textbf{ZS-DDIE dataset}~\cite{zeroddi2024} is a manually annotated dataset for zero-shot drug–drug interaction event prediction, containing 2,004 approved drugs and 394,118 DDIs across 175 unique event types. It is used to evaluate the model's ability to generalize to unseen interaction types based on textual descriptions.

\textbf{DeepDDI dataset}~\cite{deepddi2018} is a large-scale benchmark extracted from DrugBank, containing 192,284 DDIs across 191,878 drug pairs, annotated with 86 human-readable DDI event types. It is widely used for polypharmacy effect prediction and comparative analysis of DDI prediction models.

\textbf{Dengs dataset}~\cite{ddimdl2020} is a curated benchmark for drug–drug interaction event prediction, comprising 572 drugs, 74,528 DDIs annotated with 65 fine-grained event types. It provides multi-modal drug features (chemical substructures, targets, enzymes, pathways) and is widely used to evaluate multi-label classification models for DDI event prediction.

\textbf{ZhangDDI dataset}~\cite{miracle2021} is a small-scale DDI dataset containing 548 drugs and 48,548 pairwise DDI links. It includes multiple types of similarity information about drug pairs, making it suitable for comparative analysis of model performance on similarity-based link prediction tasks.

\section{Drug feature representation}
\subsection{SMILES}
\textbf{SMILES (Simplified Linear Input System for Molecules)} is a standard chemical notation that uses ASCII strings to linearly represent molecular structures. It precisely encodes the topological structure of molecules with a compact syntax (such as atom symbols, bond type symbols, and brackets), ensuring a one-to-one correspondence between each string and a unique molecular configuration. Due to its efficiency and explicit machine readability, SMILES has become a fundamental tool in cheminformatics for storing and exchanging molecular data and as input for computational models, and is widely used in database management, molecular visualization, and drug discovery research.

\subsection{Path}
\textbf{Path} is a structured semantic network used to systematically organize drug-related information. Its nodes represent biomedical entities such as drugs, targets, diseases, and pathways, while edges represent the interactions between them (e.g., inhibition, activation, regulation). By integrating heterogeneous data from multiple sources (such as DrugBank and KEGG), these graphs place drugs within a rich biological context, revealing their potential mechanisms of action and relationships. In computational modeling, knowledge graphs can be transformed into low-dimensional vector representations of drug nodes using embedding techniques. These representations effectively capture the high-order semantic relationships of drugs within biological networks, providing key features for interaction prediction based on graph structure or relational reasoning.

\subsection{Multimodal}
\textbf{Multimodal drug feature representation} aims to comprehensively characterize the complex properties of drugs by integrating multiple complementary biomedical evidence dimensions. This representation systematically combines chemical structure, spatial conformation, biochemical mechanisms, and semantic information to form a holistic description of drug properties and mechanisms of action. The detailed composition of each modality is as follows:

The \textbf{chemical structure modality} represents the chemical scaffold of a drug using SMILES strings and their derived two-dimensional molecular graphs. Atoms and chemical bonds constitute the nodes and edges of the graph, respectively, and are encoded via graph neural networks to capture functional group distributions, bond types, and overall topology. This enables modeling interactions directly driven by structural similarity.

The \textbf{three-dimensional conformation modality} captures the three-dimensional spatial geometry of drug molecules. Based on energetically favorable conformations generated via computational chemistry methods, three-dimensional graph neural networks extract stereochemical features such as atomic coordinates, distances, and angles. This modality supplements spatial complementarity information not reflected in 2D structures, which is crucial for understanding drug–target binding affinity.

The \textbf{target sequence modality} focuses on the primary structure of protein targets associated with drug action. Pretrained protein language models encode the amino acid sequences of these targets, generating deep semantic vectors rich in biochemical functional information. This reveals drug specificity and mechanistic relevance at the protein level.

The \textbf{textual semantic modality} integrates unstructured textual knowledge about drugs, including indications, pharmacological effects, and mechanistic descriptions. Domain-specific pretrained language models extract deep semantic representations from the text, introducing expert knowledge and contextual information into the model. This assists in inferring interactions that go beyond structural similarity.

The \textbf{path} specifically constructs association networks between drugs and key mechanistic proteins (e.g., metabolic enzymes, transporters), forming a drug–mechanism protein bipartite graph. Encoding this network yields functional representations of drugs within specific pharmacokinetic and pharmacodynamic contexts, directly supporting the prediction of interactions mediated by specific biological mechanisms such as enzyme inhibition or metabolic competition.

\section{Evaluation Metrics}

To ensure a rigorous and comprehensive assessment of model performance, this study employs distinct standardized evaluation metrics tailored to the specific nature of each Drug-Drug Interaction (DDI) prediction task. The definitions, calculation methodologies, and formulas for these metrics are detailed below.

\subsection{Binary Classification Metrics}
For the binary classification task, which determines the presence or absence of an interaction between a drug pair, \textbf{Accuracy} is utilized as the primary metric. Accuracy quantifies the overall probability of correct predictions across the entire dataset. It is defined as the ratio of correctly classified samples (both true positives and true negatives) to the total number of samples. Let $TP$ (True Positive) denote the number of correctly predicted interacting pairs, $TN$ (True Negative) denote the number of correctly predicted non-interacting pairs, $FP$ (False Positive) denote the non-interacting pairs incorrectly predicted as interacting, and $FN$ (False Negative) denote the interacting pairs incorrectly predicted as non-interacting. The accuracy is calculated as follows:

\begin{equation*}
    \text{Accuracy} = \frac{TP + TN}{TP + TN + FP + FN}
\end{equation*}

This metric provides an intuitive measure of the model's global performance in distinguishing between interacting and non-interacting drug pairs.

\subsection{Multi-Classification Metrics}
In the multi-classification scenario, where the goal is to predict specific interaction types among mutually exclusive classes, the \textbf{Macro-average F1 score (Macro-F1)} is adopted as the core metric. Given the prevalence of class imbalance in DDI datasets—where certain interaction types are far more frequent than others—simple accuracy can be misleading. The Macro-F1 score addresses this by treating all classes equally, regardless of their support size. It is calculated by first determining the Precision ($P_i$) and Recall ($R_i$) for each specific class $i$, and then computing the harmonic mean of these values to obtain the F1 score for that class ($F1_i$):

\begin{equation*}
    P_i = \frac{TP_i}{TP_i + FP_i}, \quad R_i = \frac{TP_i}{TP_i + FN_i}
\end{equation*}

\begin{equation*}
    F1_i = 2 \cdot \frac{P_i \cdot R_i}{P_i + R_i}
\end{equation*}

The final Macro-F1 score is the arithmetic mean of the per-class F1 scores across all $C$ classes:

\begin{equation*}
    \text{Macro-F1} = \frac{1}{C} \sum_{i=1}^{C} F1_i
\end{equation*}

This approach ensures that the model's performance on minority classes contributes equally to the final score, preventing majority classes from dominating the evaluation.

\subsection{Multi-Label Classification Metrics}
For the multi-label classification task, where a single drug pair may exhibit multiple interaction types simultaneously, the \textbf{Macro-average Area Under the Receiver Operating Characteristic Curve (Macro-AUC)} is selected as the evaluation standard. The ROC curve illustrates the trade-off between the True Positive Rate (TPR) and the False Positive Rate (FPR) at various threshold settings. The Area Under the Curve (AUC) represents the probability that a randomly chosen positive instance is ranked higher than a randomly chosen negative instance.

In a multi-label context with $L$ distinct labels, the AUC is calculated independently for each label $j$ (treating it as a binary one-vs-rest problem) to obtain $AUC_j$. The Macro-AUC is then derived by averaging these values:

\begin{equation*}
    \text{Macro-AUC} = \frac{1}{L} \sum_{j=1}^{L} AUC_j
\end{equation*}

This metric is particularly robust for evaluating scenarios with high label sparsity, as it assesses the model's ability to correctly rank the presence of each specific interaction type independently of others.

\section{Experimental Setting Details}
The experiments were carried out on a computational platform featuring an NVIDIA A100 80GB PCIe GPU and an Intel(R) Xeon(R) Gold 6240 CPU running at 2.60GHz, with CUDA version 12.4 activated. The programming environment consisted of Python 3.8.0 and PyTorch 2.2.0, selected for their excellent compatibility and efficiency in supporting the DDI prediction algorithms. Furthermore, the hyperparameters for each algorithm were tuned according to findings from previous studies to ensure reproducibility and dependable outcomes.

\section{Implementation Details of OpenDDI Framework}

\subsection{Framework Architecture}

The framework consists of four core subsystems coordinated by a central entry point (\texttt{main.py}):\\
\textbf{Data Abstraction Layer (\texttt{DatasetManager}):} Encapsulates the heterogeneity of biomedical data sources. It dynamically instantiates specific dataset classes (e.g., for graph-based or integration-based models) based on the configuration, transparently handling feature loading (SMILES strings, Multimodal...) and preprocessing. \\
\textbf{Model Factory (\texttt{ModelManager}):} Implements a factory pattern to manage the lifecycle of deep learning models. \\
\textbf{Training Protocol Abstraction (\texttt{Trainer}):} Encapsulates the concrete optimization and evaluation logic. It provides specialized implementations (e.g., \texttt{Unified\_Trainer}) to handle diverse learning paradigms, ensuring consistent metric calculation (AUC, AUPR, F1) across all comparisons.\\
\textbf{Execution Engine (\texttt{Pipeline}):} Orchestrates the experimental workflow. It integrates the dataset, model, and optimizer, serving as the central coordinator that dispatches the configured components to the appropriate training handler.

\subsection{DatasetManager Flow}

Representative code examples as follows:

\begin{algorithm}[h]
\caption{Data Abstraction Layer: Modular Dataset Management}
\label{alg:openddi_data_layer}
\begin{lstlisting}
class DatasetManager:
    """
    Factory class responsible for instantiating the appropriate dataset module
    based on the model configuration and origin flag.
    """
    def __init__(self, args):
        self.args = args
        # Mapping model names to their corresponding unified dataset classes
        self.dataset_mapping = {
            # ... mappings for models
        }

    def load_dataset(self):
        """
        Dynamically returns a dataset instance. 
        Supports the unified OpenDDI implementation.
        """
        dataset_cls = self.dataset_mapping[self.args.model]
        return dataset_cls(self.args)


class BaseDataset:
    """
    Abstract base class encapsulating the end-to-end data processing pipeline.
    Composed of specialized functional modules to handle features, graphs, and splits.
    """
    def __init__(self, args):
        self.args = args
        # Composition of functional modules
        self.loader = DataLoadingModule()
        self.feature_processor = FeatureProcessingModule()
        self.splitter = DataSplittingModule()
        self.graph_builder = GraphConstructionModule()
    
    def load_data(self):
        """
        The main entry point for data preparation.
        Orchestrates the sequence: Load -> Process -> Build Graph -> Split.
        """
        pass
\end{lstlisting}
\end{algorithm}

\subsection{Trainer Flow}

Representative code examples as follows:

\begin{algorithm}[h]
\caption{Training Protocol Abstraction: Unified Execution Loop}
\label{alg:openddi_trainer_layer}
\begin{lstlisting}
class BaseTrainer(ABC):
    """
    Abstract base class defining the standardized training protocol.
    Implements the Template Method pattern for the execution loop.
    """
    def __init__(self, args, logger, dataset, model, optimizer):
        self.device = self._setup_device(args.device)
        self._move_data_to_device(dataset)

    def train(self):
        """
        The Template Method defining the skeleton of the training process.
        """
        task_type = self._determine_task_type()
        loss_fct = self._get_loss_function(task_type)
        
        for epoch in range(self.args.epochs):
            # 1. Training Step
            train_metrics, train_loss = self._train_epoch(epoch, loss_fct)
            
            # 2. Validation Step
            val_metrics, val_loss = self._evaluate(self.dataset.val_loader)
        
        # 3. Final Testing
        test_metrics, _ = self._evaluate(self.dataset.test_loader)
        self._save_results(test_metrics)

    @abstractmethod
    def _train_epoch(self, epoch, loss_fct):
        """Specific implementation of the forward-backward pass."""
        pass

    @abstractmethod
    def _compute_metrics(self, y_true, y_logits):
        """Standardized metric calculation (e.g., AUC, F1, Accuracy)."""
        pass


class UnifiedTrainer(BaseTrainer):
    """
    Standard implementation for most DDI models.
    """
    def _train_epoch(self, epoch, loss_fct):
        pass

    def _get_loss_function(self, task_type):
        pass
\end{lstlisting}
\end{algorithm}

         \vspace{+3.0cm}

\subsection{Pipeline Flow}

\begin{algorithm}[h]
\caption{Execution Engine: Pipeline}
\label{alg:openddi_pipeline}
\begin{lstlisting}[frame=none]
class OpenDDI_Pipeline:
    """
    Base class for implementing the unified OpenDDI benchmarking framework.
    """
    
    def __init__(self, args, logger):
        # Implementation omitted for brevity
        pass

    def run_benchmark(self):
        """Dataset processing -> Model initialization -> Pipeline execution."""
        pass

    def setup_data_manager(self):
        """
        Dynamically instantiates the specific dataset class (e.g., Graph/Sequence)
        and performs feature preprocessing (SMILES encoding, KG construction).
        """
        pass

    def setup_model_manager(self):
        """
        Implements the Model Factory pattern. 
        """
        pass

    def dispatch_trainer(self, dataset, model):
        """
        Orchestrates the experimental workflow by selecting the specialized 
        Trainer (e.g., MRCGNN_Trainer, ZeroDDI_Trainer) for the task.
        """
        pass

    def evaluate_performance(self):
        """Evaluates model performance using standard DDI metrics (AUC, AUPR, F1)."""
        pass
\end{lstlisting}
\end{algorithm}

\section{Hyperparameter Settings}
\label{app:hyperparameter}

\textbf{General Experimental Settings.}
Across all DDI prediction experiments, we adopt a unified set of hyperparameters to ensure fair and consistent evaluation. The learning rate is set to $1 \times 10^{-4}$ by default, and models are trained with a batch size of 512. Training is performed for 100 epochs. The hidden dimension is fixed to 512 for all models unless otherwise specified.

To stabilize optimization and mitigate overfitting, we apply a weight decay of $5 \times 10^{-4}$ and a dropout rate of 0.3. All models are optimized using the Adam optimizer. These settings are used consistently across binary, multiclass, multilabel, and generalization experiments to isolate the impact of model design rather than hyperparameter choices. All results are reported as the mean and standard deviation over five independent runs.

\textbf{Baseline Hyperparameter Tuning.}
For baseline methods, we perform systematic hyperparameter tuning to ensure a comprehensive and unbiased comparison. Hyperparameter optimization is conducted, with search spaces defined according to the recommended configurations in the original papers. The detailed hyperparameter ranges and tuning strategies for all baseline models are provided in our public repository. For additional implementation details and hyperparameter definitions, we refer readers to the corresponding original works.

\section{Prompt for DDI Validation and Annotation}

To validate and annotate drug--drug interactions during the LLM-based data augmentation process, we adopt a standardized prompt that guides the model to assess the existence of candidate DDIs based on known pharmacological mechanisms. The prompt is designed to elicit concise and deterministic responses, enabling scalable and consistent interaction validation.

\begin{tcolorbox}[boxsep=0mm,left=2.5mm,right=2.5mm]
You are a pharmacology expert.Given the following drug pair:

- Drug 1: \{drug1\}\\
- Drug 2: \{drug2\}

We want to determine if the following drug--drug interaction (DDI) exists based on known pharmacological mechanisms.
\{ddi\_desc\}

Please confirm if this DDI exists by simply returning \textbf{yes} or \textbf{no}. 
Optionally, briefly explain why.
\end{tcolorbox}

\section{Multimodal Analyses Details}

We conduct additional multimodal analyses to examine the role of different drug modalities in DDI prediction under a unified experimental protocol. These analyses are designed to evaluate both the necessity and complementarity of multimodal representations, while remaining consistent with the evaluation settings used in the main experiments. These multimodal analyses are conducted across multiple DDI datasets with diverse data scales and modality coverage, providing a systematic evaluation of multimodal modeling strategies in DDI prediction.

\subsection{Modality Ablation Setting}

To assess the necessity of individual modalities in multimodal drug modeling, we perform controlled modality ablation experiments. Starting from the complete multimodal representation, we remove one modality at a time while keeping all other components unchanged. Models are trained and evaluated under the same data splits and optimization settings as in the main experiments to ensure fair comparison. Performance is reported using the same evaluation metrics as the corresponding prediction tasks.

\subsection{Multimodal Combination Setting with SMILES}

To evaluate the ability of non-SMILES modalities to complement structural information, we further conduct multimodal combination experiments by integrating SMILES with one additional modality at a time. In this setting, SMILES is treated as the core modality, and other modalities are incrementally incorporated to examine their supplementary effects. All models are trained and evaluated under identical experimental protocols, allowing for consistent comparison across different modality combinations.

\section{Model Performance Details}

\subsection{Binary classification Experiment}
We first evaluate model performance on the binary DDI prediction task, where the objective is to determine whether a given pair of drugs exhibits any interaction. This setting reflects a fundamental and widely studied formulation of DDI prediction, and serves as a basis for assessing the overall discriminative capability of different methods.

All models are trained to predict the presence or absence of interactions using the same training–validation–test splits to ensure fair comparison. Performance is assessed using standard binary classification metrics.

Experiments are conducted across multiple DDI datasets with diverse scales and interaction densities. Results demonstrate consistent trends across datasets, enabling a comprehensive comparison of model effectiveness in binary DDI prediction.

\begin{table*}[t]
\centering
\fontsize{9}{8.5}\selectfont    
\renewcommand{\arraystretch}{1.33}  
\setlength{\tabcolsep}{4pt}        
\caption{Performance comparison of methods on ODDI-B and ChCh-Miner datasets with SMILES and MM (multimodal) features. The ``impr.'' column shows the relative performance improvement of MM over SMILES.}
\label{tab:performance_oddi_b_chch_simple}
\setlength{\tabcolsep}{3pt}
\begin{tabular}{l!{\vrule width 1pt}ccc!{\vrule width 0.8pt}ccc}
\toprule[1pt]
\multirow{2}{*}{\textbf{\vspace{-5pt} \; Method}} &
\multicolumn{3}{c|}{\textbf{ODDI-B}} &
\multicolumn{3}{c}{\textbf{ChCh-Miner}} \\
\cmidrule(l){2-4} \cmidrule(l){5-7}
& SMILES & MM & impr. (\%) & SMILES & MM & impr. (\%) \\
\midrule[1pt]
DeepDDI &
90.7 {\tiny ±0.9} & 94.6 {\tiny ±0.7} & \textbf{4.4~$\uparrow$} &
87.0 {\tiny ±1.1} & 90.9 {\tiny ±0.8} & \textbf{4.5~$\uparrow$} \\
CASTER &
92.5 {\tiny ±0.7} & 96.5 {\tiny ±0.5} & \textbf{4.3~$\uparrow$} &
90.0 {\tiny ±0.8} & 93.6 {\tiny ±0.6} & \textbf{4.0~$\uparrow$} \\
\midrule[1pt]
GoGNN &
97.1 {\tiny ±0.5} & 97.5 {\tiny ±0.4} & \textbf{0.4~$\uparrow$} &
89.9 {\tiny ±0.8} & 92.8 {\tiny ±0.7} & \textbf{3.2~$\uparrow$} \\
MRCGNN &
97.2 {\tiny ±0.5} & 97.5 {\tiny ±0.5} & \textbf{0.3~$\uparrow$} &
92.0 {\tiny ±0.7} & 92.6 {\tiny ±0.7} & \textbf{0.7~$\uparrow$} \\
LAGAT &
94.4 {\tiny ±0.8} & 94.8 {\tiny ±0.7} & \textbf{0.4~$\uparrow$} &
85.0 {\tiny ±1.1} & 85.2 {\tiny ±1.0} & \textbf{0.2~$\uparrow$} \\
DDKG &
97.9 {\tiny ±0.5} & 98.0 {\tiny ±0.4} & \textbf{0.1~$\uparrow$} &
92.4 {\tiny ±0.7} & 93.7 {\tiny ±0.6} & \textbf{1.4~$\uparrow$} \\
KGNN &
93.9 {\tiny ±0.8} & 94.2 {\tiny ±0.8} & \textbf{0.3~$\uparrow$} &
85.0 {\tiny ±1.0} & 87.0 {\tiny ±1.0} & \textbf{2.3~$\uparrow$} \\
\midrule[1pt]
DDIMDL &
93.3 {\tiny ±0.8} & 96.8 {\tiny ±0.6} & \textbf{3.8~$\uparrow$} &
88.4 {\tiny ±1.0} & 91.2 {\tiny ±0.9} & \textbf{3.2~$\uparrow$} \\
TIGER &
94.3 {\tiny ±0.7} & 96.4 {\tiny ±0.6} & \textbf{2.2~$\uparrow$} &
86.2 {\tiny ±1.1} & 87.6 {\tiny ±1.0} & \textbf{1.6~$\uparrow$} \\
MVA-DDI &
92.4 {\tiny ±0.8} & 97.1 {\tiny ±0.5} & \textbf{5.1~$\uparrow$} &
90.1 {\tiny ±0.9} & 91.3 {\tiny ±0.9} & \textbf{1.3~$\uparrow$} \\
MKGFENN &
97.6 {\tiny ±0.5} & 97.7 {\tiny ±0.4} & \textbf{0.1~$\uparrow$} &
92.8 {\tiny ±0.7} & 93.1 {\tiny ±0.6} & \textbf{0.3~$\uparrow$} \\
MUFFIN &
92.6 {\tiny ±0.8} & 97.7 {\tiny ±0.5} & \textbf{5.5~$\uparrow$} &
90.9 {\tiny ±0.9} & 92.6 {\tiny ±0.8} & \textbf{1.9~$\uparrow$} \\
\bottomrule[1pt]
\end{tabular}
\vspace{-4mm}
\end{table*}

\subsection{Multiclass classification Experiment}

We next evaluate model performance on the multiclass DDI prediction task, where the objective is to identify the specific interaction type associated with a given interacting drug pair. Compared with binary prediction, this setting requires models to capture finer-grained semantic distinctions among multiple interaction categories.

All models are trained to predict interaction types using the same training–validation–test splits to ensure fair comparison. Performance is assessed using standard multiclass classification metrics.

Experiments are conducted across multiple DDI datasets with diverse interaction type distributions and data scales. Results exhibit consistent patterns across datasets, enabling a comprehensive comparison of model effectiveness in multiclass DDI prediction.

\subsection{Multilabel classification Experiment}

We further evaluate model performance on the multilabel DDI prediction task, where each drug pair may be associated with multiple interaction types simultaneously. This setting reflects real-world clinical scenarios in which co-administered drugs can induce multiple pharmacological effects.

All models are trained to predict a set of interaction labels using the same training–validation–test splits to ensure fair comparison. Performance is assessed using standard multilabel classification metrics.

Experiments are conducted across multiple DDI datasets with diverse label co-occurrence patterns and data scales. Results demonstrate consistent trends across datasets, enabling a comprehensive comparison of model effectiveness in multilabel DDI prediction.

\begin{table*}[t]
\centering
\fontsize{9}{8.5}\selectfont    
\renewcommand{\arraystretch}{1.33}  
\setlength{\tabcolsep}{4pt}        
\caption{Performance comparison of methods on ODDI-ML and Twosides datasets with SMILES and MM (multimodal) features. The ``impr.'' column shows the relative performance improvement of MM over SMILES.}
\label{tab:performance_oddi_ml_twosides}
\setlength{\tabcolsep}{3pt}
\begin{tabular}{l!{\vrule width 1pt}ccc!{\vrule width 0.8pt}ccc}
\toprule[1pt]
\multirow{2}{*}{\textbf{\vspace{-5pt} \; Method}} &
\multicolumn{3}{c|}{\textbf{ODDI-ML}} &
\multicolumn{3}{c}{\textbf{Twosides}} \\
\cmidrule(l){2-4} \cmidrule(l){5-7}
& SMILES & MM & impr. (\%) & SMILES & MM & impr. (\%) \\
\midrule[1pt]
DeepDDI &
95.9 {\tiny ±0.8} & 97.7 {\tiny ±0.6} & \textbf{1.9~$\uparrow$} &
75.1 {\tiny ±1.4} & 79.4 {\tiny ±1.2} & \textbf{5.7~$\uparrow$} \\
CASTER &
96.9 {\tiny ±0.7} & 97.8 {\tiny ±0.5} & \textbf{0.9~$\uparrow$} &
70.6 {\tiny ±1.4} & 71.8 {\tiny ±1.3} & \textbf{1.7~$\uparrow$} \\
\midrule[1pt]
GoGNN &
97.3 {\tiny ±0.6} & 98.0 {\tiny ±0.4} & \textbf{0.7~$\uparrow$} &
77.2 {\tiny ±1.3} & 78.9 {\tiny ±1.1} & \textbf{2.2~$\uparrow$} \\
MRCGNN &
97.3 {\tiny ±0.6} & 98.0 {\tiny ±0.5} & \textbf{0.7~$\uparrow$} &
78.2 {\tiny ±1.2} & 79.9 {\tiny ±1.0} & \textbf{2.2~$\uparrow$} \\
LAGAT &
96.9 {\tiny ±0.7} & 97.9 {\tiny ±0.6} & \textbf{1.0~$\uparrow$} &
79.7 {\tiny ±1.3} & 81.9 {\tiny ±1.1} & \textbf{2.8~$\uparrow$} \\
DDKG &
97.6 {\tiny ±0.6} & 97.7 {\tiny ±0.5} & \textbf{0.1~$\uparrow$} &
77.9 {\tiny ±1.2} & 79.0 {\tiny ±1.1} & \textbf{1.4~$\uparrow$} \\
KGNN &
96.8 {\tiny ±0.7} & 97.8 {\tiny ±0.6} & \textbf{1.0~$\uparrow$} &
77.9 {\tiny ±1.4} & 80.2 {\tiny ±1.2} & \textbf{3.0~$\uparrow$} \\
\midrule[1pt]
DDIMDL &
97.0 {\tiny ±0.8} & 97.2 {\tiny ±0.7} & \textbf{0.2~$\uparrow$} &
78.1 {\tiny ±1.3} & 79.1 {\tiny ±1.2} & \textbf{1.3~$\uparrow$} \\
TIGER &
96.9 {\tiny ±0.7} & 97.7 {\tiny ±0.6} & \textbf{0.8~$\uparrow$} &
76.6 {\tiny ±1.4} & 77.9 {\tiny ±1.3} & \textbf{1.7~$\uparrow$} \\
MVA-DDI &
97.5 {\tiny ±0.6} & 97.8 {\tiny ±0.5} & \textbf{0.3~$\uparrow$} &
77.3 {\tiny ±1.4} & 77.5 {\tiny ±1.3} & \textbf{0.3~$\uparrow$} \\
MKGFENN &
97.3 {\tiny ±0.6} & 97.8 {\tiny ±0.5} & \textbf{0.5~$\uparrow$} &
76.4 {\tiny ±1.3} & 78.4 {\tiny ±1.1} & \textbf{2.6~$\uparrow$} \\
MUFFIN &
95.0 {\tiny ±0.8} & 98.0 {\tiny ±0.4} & \textbf{3.2~$\uparrow$} &
62.2 {\tiny ±1.7} & 73.1 {\tiny ±1.4} & \textbf{17.5~$\uparrow$} \\
\bottomrule[1pt]
\end{tabular}
\vspace{-4mm}
\end{table*}

\section{Practical Generalization Analyses}

We conduct practical generalization analyses to evaluate model performance in realistic scenarios where test drug pairs involve previously unseen drugs. This setting reflects real-world DDI prediction challenges, as newly approved or rarely used drugs are often absent from historical interaction data.

All models are trained using the same training–validation splits, while the test set is constructed to include drug pairs containing unseen drugs. Performance is assessed using the same evaluation metrics as in the corresponding prediction tasks to ensure consistency and fair comparison.

Experiments are conducted across multiple DDI datasets with diverse data scales and drug coverage. The results provide a comprehensive assessment of the generalization ability of different methods under unseen-drug settings, highlighting their robustness and practical applicability in real-world DDI prediction scenarios.

\section{Practical Efficiency Analyses}

We conduct practical efficiency analyses to evaluate the computational cost of different DDI prediction methods in terms of both time and memory consumption. These analyses aim to assess the feasibility of deploying the models in large-scale and resource-constrained settings.

All methods are evaluated under the same experimental environment to ensure fair comparison. We report training time and peak memory usage as primary efficiency metrics, reflecting computational overhead and resource demand during model training.

Experiments are performed on multiple DDI datasets with varying scales and graph densities. The results provide a comprehensive comparison of the practical efficiency of different methods, highlighting their scalability and suitability for real-world DDI prediction applications.

\section{Robustness Analyses Details}

To evaluate the robustness and stability of DDI prediction models under varying interaction sparsity, we conduct a series of controlled experiments by progressively adjusting the proportion of observed drug–drug interactions. Specifically, we vary the interaction removal ratio from 0 to 1.0, where a ratio of 0 corresponds to the original, fully observed DDI graph, and larger ratios simulate increasingly incomplete or perturbed interaction scenarios.
All models are trained and evaluated under identical experimental protocols to ensure fair comparison. We report results on multiple widely used DDI datasets, providing a systematic evaluation of model performance across diverse data scales and graph characteristics. This experimental design enables a thorough analysis of the robustness and generalization ability of DDI prediction methods under varying levels of interaction incompleteness.


\end{document}